\documentclass{svjour3}
\usepackage{pgf}
\usepackage{tikz}
\usetikzlibrary{calc,arrows,matrix,graphs,shapes,snakes,automata,backgrounds,petri,positioning,decorations.pathmorphing}
\usepackage{verbatim}
\usepackage{amsmath}
\usepackage{amssymb}
\usepackage{latexsym}
\usepackage{proof}
\usepackage{array}
\usepackage{calc}
\usepackage{mathdots}
\usepackage{stmaryrd}
\usepackage[numbers,sort&compress]{natbib}
\usepackage[english]{babel}
\usepackage{linguex}
\usepackage{extarrows}
\usepackage{bussproofs}
\usepackage{xcolor}
\usepackage{rotating}
\usepackage[colorlinks,citecolor=red]{hyperref}

\newcommand{\symon}[1]{}
\newcommand{\moot}[1]{}

\newcommand{\editout}[1]{}

\newcommand{\str}{st}
\newcommand{\ldl}{\mathbin{\backslash}}
\newcommand{\ldr}{\mathbin{/}}

\newcommand{\himpl}{\mathbin{\multimap}}
\newcommand{\bridge}{\,\hat{\,}}
\newcommand{\bo}{[}
\newcommand{\bc}{]}

\newcommand{\proofspace}{\vphantom{()}}

\newcommand{\seq}{\Rightarrow\proofspace}

\newcommand{\nd}{\vdash\proofspace}

\newcommand{\apsnode}[2]{\overset{#1}{\underset{#2^{\rule{0pt}{1.2ex}}}{\centerdot}}}

\newcommand{\apsnodei}{\centerdot_{\rule{0pt}{1.5ex}}}
\newcommand{\nodeindex}[1]{\textcolor{white}{\textbf{#1}}}

\tikzset{pas/.style={fill=gray!60}, 
act/.style={fill=gray!30},
main/.style={draw,fill=white},
ctx/.style={rounded rectangle,minimum size=7mm},
val/.style={rectangle,minimum size=7mm},
cmd/.style={chamfered rectangle,draw,fill=white},
tns/.style={circle,minimum size=4.5mm,draw,fill=white},  
par/.style={circle,minimum size=4.5mm,draw,fill=black},  
ttns/.style={circle,minimum size=4.5mm,inner sep=2pt,draw,fill=white,double=white},  
ppar/.style={circle,minimum size=4.5mm,inner sep=2pt,draw=white,fill=black,double=black},  
minipar/.style={circle,minimum size=5mm,draw,fill=black}, 
pn/.style={rounded corners, rectangle,fill=gray!30,draw,minimum size=15mm},
medpn/.style={rounded corners, rectangle,fill=gray!30,draw,minimum size=20mm},
 bigpn/.style={rounded corners, rectangle,fill=gray!30,draw,minimum size=25mm}}

\tikzset{every picture/.style={scale=.9, transform shape, node distance=7mm}}

\newcommand{\qedsym}{\hfill $\Box$}

\newenvironment{scprooftree}[1]%
  {\gdef\scalefactor{#1}\begin{center}\proofSkipAmount \leavevmode}%
  {\scalebox{\scalefactor}{\DisplayProof}\proofSkipAmount \end{center} }

\begin{document}

\journalname{HAL/arXiv}
\title{Logical foundations for hybrid type-logical grammars\thanks{This is a significantly extended version of a paper presented at Formal Grammar 2019 \citep{msg19htlg}. We thank the referees of Formal Grammar as well as the audience of the conference for their invaluable feedback.}}
\author{Richard Moot and Symon Jory Stevens-Guille}
\institute{LIRMM, Universit\'{e} de Montpellier, CNRS \and The Ohio State University}

\maketitle

\begin{abstract}
This paper explores proof-theoretic aspects of hybrid type-logical grammars, a logic combining Lambek grammars with lambda grammars. We prove some basic properties of the calculus, such as normalisation and the subformula property and also present both a sequent and a proof net calculus for hybrid type-logical grammars. In addition to clarifying the logical foundations of hybrid type-logical grammars, the current study opens the way to variants and extensions of the original system, including but not li\-mi\-ted to a non-associative version and a multimodal version incorporating structural rules and unary modes.
\end{abstract}

\section{Introduction}

Hybrid type-logical grammar (HTLG) is a logic introduced by \citet{kl12gap}. The logic combines the standard Lambek grammar implications with the lambda grammar operations. As a consequence, the lambda calculus term constructors of abstraction and application live side-by-side with the Lambek calculus operation of concatenation and its residuals. The  logic is motivated by empirical limitations of its subsystems. It provides a simple account of many phenomena on the syntax-semantics interface, for which neither of its subsystems has equally simple solutions \cite{kl12gap,kl13dgap,kl15lp,kl13coord,kl20tls}. 

For instance, Lambek calculi struggle to account for medial extraction, as is required for the wide-scope reading of the universal in (\ref{1}). Such cases are straightforwardly accounted for by lambda grammars. Lambda grammars treat verbs missing subjects and verbs missing objects as having the same linear syntactic type, e.g.\ \textit{Cthulhu devoured} and \textit{devoured Cthulhu} are both just sentences missing a noun phrase somewhere and so can be coordinated \cite{worth14coord,k10,kl15lp}. For the same reason --- namely the absence of directionality --- lambda grammars cannot easily distinguish (\ref{2}) from (\ref{3}), whereas the distinction is trivial to implement in Lambek calculi. 

\begin{enumerate}
	\item \label{1} Someone delivers everything to its destination.  
	\item \label{2} Ahmed loves and the pizza dislikes Johani.
	\item \label{3} Ahmed loves and Johani dislikes the pizza.
\end{enumerate} 

In their paper on determiner gapping in hybrid type-logical grammar, \citet[footnote 7]{kl13dgap} `acknowledge that there remains an important theoretical issue: the formal properties of our hybrid implicational logic are currently unknown'.

The rest of this paper is structured as follows. In Section~\ref{sec:nd}, we present the natural deduction calculus of \citet{kl12gap} and prove some basic properties of the calculus, namely normalisation, decidability and the subformula property. In Section~\ref{sec:seq}, we present a sequent calculus for HTLG, prove cut elimination for it, and prove that it is equivalent to the natural deduction calculus. In Section~\ref{sec:pn} we present a proof net calculus for the HTLG, prove it is correct and give a cut elimination proof for the calculus. In Section~\ref{sec:compl}, we conclude with an analysis of the complexity of different versions of HTLG.

Taken together, the results in this paper put HTLG on a firm theoretical foundation. It also provides a framework for extensions of the logic, showing that adding structural rules as used in \cite{kl20tls} does not pose a theoretical problem. As a final application of the results in this paper, the proof net calculus provides a proof search method which is both flexible and transparent.


\section{Natural deduction}
\label{sec:nd}

The basic syntactic objects of HTLG are tuples, where the first element is a linear lambda term and the second is a type-logical formula drawn from the union of implicational linear logic and Lambek formulas. Given a set of atomic formulas $A$ (we will assume $A$ contains at least the atomic formula $n$ for noun, $np$ for noun phrase, $s$ for sentence, and $pp$ for prepositional phrase), the formula language of HTLG is the following.
\begin{itemize}
	\item $T_{Lambek} ::=  A \;\;\vert\;\; T_{Lambek} \slash T_{Lambek} \;\; \vert \;\;  T_{Lambek} \backslash T_{Lambek}$
	\item $T_{Logic}$ ::= $T_{Lambek}$ $\;\vert\;$ $T_{Logic}  \multimap T_{Logic}$
\end{itemize}

Prosodic types are simple types with a unique atomic type $\str$ (for \emph{structure} or, in an associative context, \emph{string}). Logical formulas are translated to prosodic types as follows.
\begin{align*}
\textit{Pros}(T_{\textrm{Lambek}}) & = \str \\
\textit{Pros}(T_{\textrm{Logic}}  \multimap T_{\textrm{Logic}}) & = \textit{Pros}(T_{\textrm{Logic}}) \rightarrow  \textit{Pros}(T_{\textrm{Logic}})
\end{align*}


The lambda terms of HTLG, called \emph{prosodic terms}, are constructed as follows.
\begin{itemize}
	\item Atoms: $+^{\str\rightarrow \str \rightarrow \str}$, $\epsilon^{\str}$, a countably infinite number of variables $x_0,x_1,\ldots$ for each type $\alpha$; by convention we use $p, q, \ldots$ for variables of type $\str$.
	\item Construction rules:
\begin{itemize}
	\item   if $M^{\alpha\rightarrow\beta}$ and $N^{\alpha}$, then  $(MN)^\beta$
	\item if $x^\alpha$ and $M^{\beta}$, then $(\lambda x.M)^{\alpha \rightarrow \beta}$
\end{itemize}
\end{itemize}

In what follows, we restrict the prosodic terms to linear lambda terms, requiring each $\lambda$ binder to bind exactly one occurrence of its variable $x$. This restriction is standard in HTLG.



\begin{figure}
	\centering
	\begin{center}
	\AxiomC{}
	\RightLabel{\textit{Lex}}
	\UnaryInfC{\ensuremath{p^{\str}:w \vdash M:A}}	
	\DisplayProof
	\hskip 1.5em
		\AxiomC{}
	\RightLabel{\textit{Ax}}
	\UnaryInfC{\ensuremath{x^{\alpha}:A \vdash x^{\alpha}:A}}	
	\DisplayProof
	\end{center}
	\begin{center}
		\AxiomC{\ensuremath{\Gamma \vdash N^\alpha:A}}
		\AxiomC{\ensuremath{\Delta \vdash M^{\alpha \rightarrow \beta}:A \multimap B}}
		\RightLabel{\ensuremath{\multimap E}}
		\BinaryInfC{\ensuremath{\Gamma, \Delta \vdash (MN)^\beta:B}}
		\DisplayProof
		\hskip 1.5em
		\AxiomC{\ensuremath{\Gamma, x^\alpha:A \vdash M^\beta:B}}
		\RightLabel{\ensuremath{\multimap I}}
		\UnaryInfC{\ensuremath{\Gamma \vdash (\lambda x.M)^{\alpha\rightarrow\beta}:A \multimap B}}
		\DisplayProof
	\end{center}
	\begin{center}
		\AxiomC{\ensuremath{\Gamma \vdash M^{\str}:A\slash B}}
		\AxiomC{\ensuremath{\Delta \vdash N^{\str}:B}}
		\RightLabel{\ensuremath{\slash E}}
		\BinaryInfC{\ensuremath{\Gamma, \Delta \vdash (M+N)^{\str}:A}}
		\DisplayProof
		\hskip 1.5em
		\AxiomC{\ensuremath{\Gamma, p^{\str}:A \vdash (M+p)^{\str}:B}}
		\RightLabel{\ensuremath{\slash I}}
		\UnaryInfC{\ensuremath{\Gamma \vdash M^{\str}:B\slash A}}
		\DisplayProof
	\end{center}
	\begin{center}
		\AxiomC{\ensuremath{\Delta \vdash M^{\str}:B}}
		\AxiomC{\ensuremath{\Gamma \vdash N^{\str}:B\backslash A}}
		\RightLabel{\ensuremath{\backslash E}}
		\BinaryInfC{\ensuremath{\Delta, \Gamma \vdash (M+N)^{\str}:A}}
		\DisplayProof
		\hskip 1.5em
		\AxiomC{\ensuremath{p^{\str}:A, \Gamma \vdash (p+M)^{\str}:B}}
		\RightLabel{\ensuremath{\backslash I}}
		\UnaryInfC{\ensuremath{\Gamma \vdash M^{\str}:A\backslash B}}
		\DisplayProof
	\end{center}
	\begin{center}
		\vskip 0.5em
	\AxiomC{\ensuremath{\Gamma\vdash M:C}}
	\RightLabel{\ensuremath{[\beta\eta]}}
	\UnaryInfC{\ensuremath{\Gamma\vdash M':C}}
	\DisplayProof
	
	\end{center}

	\caption{Gentzen-Style ND Inference Rules for HTLG}
	\label{fig:nd}
\end{figure}





The natural deduction rules for HTLG are given by Figure \ref{fig:nd}. The lexicon rule $\textit{Lex}$ assigns the word $p$ a formula $A$ and a linear lambda term $M$ of type $\textit{Pros}(A)$. Since this term $M$ is linear, it contains exactly one free occurrence of $p$. When no confusion is possible (for example when a word appears several times in a sentence), we use the word itself instead of the unique variable $p$ (the formula $w$ has a purely technical role and cannot appear on the right hand side of the lexicon or axiom rule). An example would be $\lambda P. (P \textit{everyone}):(np \multimap s)\multimap s$.

The axiom rule $\textit{Ax}$ similarly requires that $\alpha = \textit{Pros}(A)$. It is then easily verified that the other rules ensure that the term assigned to the conclusion formula $C$ is of type $\textit{Pros}(C)$ given that the rule premisses are well-typed.

The elimination rules have the standard condition that no free variables are shared between $\Gamma$ and $\Delta$, which ensures $\Gamma,\Delta$ is a valid context. The introduction rules have the standard side-condition that $\Gamma$ contains at least one formula, thereby ensuring that provable statements cannot have empty antecedents. As is usual, the premiss of the elimination rule containing the eliminated connective (the rightmost premiss for the $\himpl\textit{E}$ and $\backslash\textit{E}$ rules, and the leftmost premiss for the $/\textit{E}$ rule) is called the \emph{major premiss} of the rule and the other premiss the \emph{minor premiss}.

The $\beta\eta$ rule has the side condition that $M$ and $M'$ are $\beta\eta$ equivalent in the simply typed lambda calculus. We generally restrict applications of this rule either to 1.\ a single beta reduction step, or  2.\ cases where $M'$ is the unique long normal form of $M$. In practice, this means we can use on-the-fly normalisation of the lambda terms. 

As is standard in lambda calculus, we use $M[N]$ for a term $M$ with a unique distinguished occurrence of a subterm $N$. We similarly use $M[N][P]$ for a term $M$ with two distinguished (non-overlapping) term occurrences $N$ and $P$, and $M[N[P]]$ for the occurrence of a term $P$ inside a term $N$, itself inside a bigger term $M$.

It is often convenient to use a Prawitz-style presentation of natural deduction proofs. Prawitz-style proofs save on horizontal space by leaving the antecedent formulas (on the left-hand side of the turnstile) implicit and removing the turnstyle. Only the succedent formula (on the right-hand side of the turnstile) and its associated term are displayed. We can recover the antecedent formulas by identifying the free variables in the term.

Given the foregoing we're now in a position to give Prawitz-style natural deduction proofs in HTLG of sentences (\ref{1}) and (\ref{3}) in Figures \ref{proof1} and \ref{proof3}. The wide scope parse of sentence (\ref{1}) in Figure (\ref{proof1}) cannot be carried out in classical Lambek calculus. This is because extraction from medial positions is prevented by the introduction rules, which require hypotheses to be at the left (resp. right) edge of the context. The proof of sentence (\ref{2}), erroneously generated by the nondirectional subsystem of HTLG, is shown in Figure \ref{proof2}.\footnote{Linear Categorial Grammar \cite{mp10,p13,s10,m12,m13,worth14coord} is the restriction of HTLG to the linear implication rules. \symon{In fact, LCG differs from HTLG in that in LCG the pheno types cannot be recovered as a function of the tecto types. The proof in Figure \ref{proof2} employs a variant of the $\multimap$ rules of HTLG where the condition that $\alpha$ = \textit{Pros(A)} is not in force.}\moot{I don't understand you previous remark. I've corrected some minor mistakes in the proof, and it seems that everything is well-typed, as it should be!}\symon{When I wrote LCG's pheno as a function from the tecto in my QP, Carl wrote to me ` LCG does not work this way; there is not a functor from tectos to pheno types as there is in ACG or HTLCG. For example, English finite VPs have phenotype $s \rightarrow s$, but nonfinite VPs havephenotype $s$.'} LCG has been criticized  \cite{muskens01lfg,kl15lp} for licensing proofs parallel to the one in Figure \ref{proof2}. Note that given the types (with $X=np\multimap s$) \emph{all} lexical term assignments for this type overgenerate \cite{hal-00996724}, so this problem is not easily fixed.} The proof in Figure \ref{3} makes use of nothing but the standard Lambek connectives.\footnote{Conjunction in HTLG is an axiom schema, which we notate by means of X, where X is any HTLG tecto type. Conjunction in LCG is also an axiom schema but the variant of conjunction used in Figure \ref{proof2} is taken from \cite{kl15lp}. This version of the conjunction explicitly feeds the first conjunct an empty string to ensure the `raised' element in Right Node Raising appears adjacent to the right conjunct. One could also imagine the conjunction introducing empty strings to both conjuncts and separately appending the `raised' element to the right edge of the conjunction. This would license the following sentence, which is clearly ungrammatical:
  \begin{enumerate}
    \setcounter{enumi}{3}
\item *Ahmed loves and dislikes dessert the pizza.
\end{enumerate}
Evidently the conjunction lexeme cannot be modified to block erroneous coordinations. Observe that if the linear types in Figure \ref{proof2} are replaced with Lambek slashes, the proof is correctly predicted to fail. That LCG overgenerates with respect to coordination is a strong empirical motivation for type-logics like HTLG.}

\begin{figure}[]
    \centering
    \begin{scprooftree}{0.4}
    \AxiomC{}
    \RightLabel{Lex}
    \UnaryInfC{\ensuremath{\lambda P.P(\textit{everything}):(np\multimap s) \multimap s}}
    
    \AxiomC{}
    \RightLabel{Lex}
    \UnaryInfC{\ensuremath{\lambda P.P(\textit{someone}):(np\multimap s) \multimap s}}
    
    \AxiomC{}
    \RightLabel{Ax}
    \UnaryInfC{\ensuremath{y^{\str}:np}}
    
    \AxiomC{}
    \RightLabel{Lex}
    \UnaryInfC{\ensuremath{\textit{delivers}: ((np\backslash s)/np)/np}}
    
    \AxiomC{}
    \RightLabel{Ax}
    \UnaryInfC{\ensuremath{x^{\str}:np}}
    
    \RightLabel{/E}
    \BinaryInfC{\ensuremath{\textit{delivers} + x:((np\backslash s)/np)}}
    
    \AxiomC{}
    \RightLabel{Lex}
    \UnaryInfC{\ensuremath{\textit{to}:pp\backslash np}}
    
    \AxiomC{}
    \RightLabel{Lex}
    \UnaryInfC{\ensuremath{\textit{its}:np/n}}
    
    \AxiomC{}
    \RightLabel{Lex}
    \UnaryInfC{\ensuremath{\textit{destination}:n}}
    \RightLabel{/E}
    \BinaryInfC{\ensuremath{\textit{its} + \textit{destination}:np}}
    \RightLabel{/E}
    \BinaryInfC{\ensuremath{\textit{to} + \textit{its} + \textit{destination}:pp}}
    \RightLabel{/E}
    \BinaryInfC{\ensuremath{\textit{delivers}+x+\textit{to}+\textit{its}+\textit{destination}:np\backslash s}}
    \RightLabel{$\backslash$ E}
    \BinaryInfC{\ensuremath{y+\textit{delivers}+x+\textit{to}+\textit{its}+\textit{destination}:np\backslash s}}
    \RightLabel{$\multimap$I}
    \UnaryInfC{\ensuremath{\lambda y.y+\textit{delivers}+x+\textit{to}+\textit{its}+\textit{destination}:np\multimap s}}
    \RightLabel{$\multimap$E}
    \BinaryInfC{\ensuremath{ \textit{someone}+\textit{delivers}+x+\textit{to}+\textit{its}+\textit{destination}:s}}
    \RightLabel{$\multimap$I}
    \UnaryInfC{\ensuremath{\lambda x.\textit{someone}+\textit{delivers}+x+\textit{to}+\textit{its}+\textit{destination}:np\multimap s}}
    \RightLabel{$\multimap$E}
    \BinaryInfC{\ensuremath{\textit{someone}+\textit{delivers}+\textit{everything}+\textit{to}+\textit{its}+\textit{destination}:s}}
    \end{scprooftree}{}
    \caption{HTLG proof of Sentence 1}
    \label{proof1}
\end{figure}{}

\begin{figure}
    \centering
    \begin{scprooftree}{0.4}
    \AxiomC{}
    \RightLabel{Lex}
    \UnaryInfC{\ensuremath{\textit{Ahmed}:np}}
    
    \AxiomC{}
    \RightLabel{Lex}
    \UnaryInfC{\ensuremath{\lambda x\lambda y.y+\textit{loves}+x:np\multimap (np\multimap s)}}
    
    \AxiomC{}
    \RightLabel{Ax}
    \UnaryInfC{\ensuremath{x^{\str}:np}}
    
    \RightLabel{$\multimap$E}
    \BinaryInfC{\ensuremath{\lambda y.y+\textit{loves}+x:np\multimap s}}
    \RightLabel{$\multimap$E}
    \BinaryInfC{\ensuremath{\textit{Ahmed}+\textit{loves}+x:s}}
    \RightLabel{$\multimap$I}
    \UnaryInfC{\ensuremath{\lambda x.\textit{Ahmed}+\textit{loves}+x:np\multimap s}}
    
    \AxiomC{}
    \RightLabel{Lex}
    \UnaryInfC{\ensuremath{\lambda Q\lambda P \lambda y.P(\epsilon)+and+Q(y):X \multimap (X \multimap X)}}

    \AxiomC{}
    \RightLabel{Ax}
    \UnaryInfC{\ensuremath{z^{\str}:np}}

    \AxiomC{}
    \RightLabel{Lex}
    \UnaryInfC{\ensuremath{\lambda y\lambda x.\textit{dislikes}:np \multimap (np \multimap s)}}
    
    \AxiomC{}
    \RightLabel{Lex}
    \UnaryInfC{\ensuremath{\textit{Johani}:np}}

    \RightLabel{$\multimap$E}
    \BinaryInfC{\ensuremath{\lambda x.x+\textit{dislikes}+\textit{Johani}:np\multimap s}}
    \RightLabel{$\backslash$E}
    \BinaryInfC{\ensuremath{z+\textit{dislikes}+\textit{Johani}:s}}
    \RightLabel{$\multimap$I}
    \UnaryInfC{\ensuremath{\lambda z.z+\textit{dislikes}+\textit{Johani}:np\multimap s}}
    
    \RightLabel{$\multimap$E}
    \BinaryInfC{\ensuremath{\lambda P\lambda y.P(\epsilon)+and+y+\textit{dislikes}+\textit{Johani}:(np\multimap s)\multimap(np\multimap s)}}
    
    \RightLabel{$\multimap$E}
    \BinaryInfC{\ensuremath{\lambda y.\textit{Ahmed}+\textit{loves}+and+y+\textit{dislikes}+\textit{Johani}:np\multimap s}}
    
    \AxiomC{}
    \RightLabel{Lex}
    \UnaryInfC{\ensuremath{\lambda x.the + x:n\multimap np}}
    
    \AxiomC{}
    \RightLabel{Lex}
    \UnaryInfC{\ensuremath{pizza:n}}

    \RightLabel{$\multimap$E}
    \BinaryInfC{\ensuremath{the+pizza:np}}
    
    \RightLabel{$\multimap$E}
    \BinaryInfC{\ensuremath{\textit{Ahmed}+\textit{loves}+and+the+pizza+\textit{dislikes}+\textit{Johani}:s}}
    \end{scprooftree}{}
    \caption{Linear/ACG proof of an incorrect reading for Sentence 2}
    \label{proof2}
\end{figure}{}

\begin{figure}
    \centering
    \begin{scprooftree}{0.4}
    \AxiomC{}
    \RightLabel{Lex}
    \UnaryInfC{\ensuremath{\textit{Ahmed}np}}
    
    \AxiomC{}
    \RightLabel{Lex}
    \UnaryInfC{\ensuremath{\textit{loves}:(np\backslash s)/np}}
    
    \AxiomC{}
    \RightLabel{Ax}
    \UnaryInfC{\ensuremath{x^{\str}:np}}
    
    \RightLabel{$/$E}
    \BinaryInfC{\ensuremath{\textit{loves}+x:np\backslash s}}
    \RightLabel{$\backslash$E}
    \BinaryInfC{\ensuremath{\textit{Ahmed}+\textit{loves}+x:s}}
    \RightLabel{$/$I}
    \UnaryInfC{\ensuremath{\textit{Ahmed}+\textit{loves}:s/np}}
    
    \AxiomC{}
    \RightLabel{Lex}
    \UnaryInfC{\ensuremath{\textit{and}:(X\backslash X)/X}}
    
    \AxiomC{}
    \RightLabel{Lex}
    \UnaryInfC{\ensuremath{\textit{Johani}:np}}
    
    \AxiomC{}
    \RightLabel{Lex}
    \UnaryInfC{\ensuremath{\textit{dislikes}:(np\backslash s)/np}}
    
    \AxiomC{}
    \RightLabel{Ax}
    \UnaryInfC{\ensuremath{x^{\str}:np}}
    
    \RightLabel{$/$E}
    \BinaryInfC{\ensuremath{\textit{dislikes}+x:np\backslash s}}
    \RightLabel{$\backslash$E}
    \BinaryInfC{\ensuremath{\textit{Johani}+\textit{dislikes}+x:s}}
    \RightLabel{$/$I}
    \UnaryInfC{\ensuremath{\textit{Johani}+\textit{dislikes}:s/np}}
    
    \RightLabel{$/$E}
    \BinaryInfC{\ensuremath{\textit{and}+\textit{Johani}+\textit{dislikes}:(s/np)\backslash(s/np)}}
    
    \RightLabel{$/$E}
    \BinaryInfC{\ensuremath{\textit{Ahmed}+\textit{loves}+\textit{and}+\textit{Johani}+\textit{dislikes}:s/np}}
    
    \AxiomC{}
    \RightLabel{Lex}
    \UnaryInfC{\ensuremath{\textit{the}:np/n}}
    
    \AxiomC{}
    \RightLabel{Lex}
    \UnaryInfC{\ensuremath{\textit{pizza}:n}}

    \RightLabel{$/$E}
    \BinaryInfC{\ensuremath{\textit{the}+\textit{pizza}:np}}
    
    \RightLabel{$/$E}
    \BinaryInfC{\ensuremath{\textit{Ahmed}+\textit{loves}+\textit{and}+\textit{Johani}+\textit{dislikes}+\textit{the}+\textit{pizza}:s}}
    \end{scprooftree}{}
    \caption{Lambek grammar/HTLG proof of Sentence 3}
    \label{proof3}
\end{figure}{}

Before showing normalisation, we first prove a standard substitution lemma.

\begin{lemma}\label{lem:subst} Let $\delta_1$ be a proof of $\Gamma\vdash N:A$ and $\delta_2$ a proof of $\Delta,x:A\vdash M[x]:C$ such that $N$ and $M$ share no free variables, then there is a proof of $\Gamma,\Delta\vdash M[N]:C$.
\end{lemma}

\paragraph{Proof} We can combine the two proofs as follows, replacing the hypothesis $x:A$ of $\delta_2$ by the proof $\delta_1$.
\[
\infer*[\delta_2\bo x:=N\bc]{\Gamma,\Delta\nd M[N]:C}{\infer*[\delta_1]{\Gamma\nd N:A}{}}
\]
Given that, by construction, $M$ and $N$ share no free variables, replacing $x$ by $N$  cannot make a rule application in $\delta_2$ invalid. \qedsym

Given that the $w$ atomic formula appearing on the left-hand side of the \textit{Lex} rule is by construction forbidden to appear on the right-hand side of a sequent, this means that the substitution lemma can never apply to a lexical hypothesis (since there are no proofs of the form $\Gamma\vdash N:w$). 


	

\subsection{Normalisation}

We show that HTLG is normalising. A normal form for an HTLG proof is defined as follows.

\begin{definition}
	A derivation $D$ for HTLG is \emph{normal} iff each major premiss of an elimination rule is either:
	\begin{enumerate}
		\item an assumption
		\item a conclusion of an application of an E-rule. 
	\end{enumerate} 
\end{definition}

In general, we call a logic \emph{normalising} just in case there is an effective procedure for extracting normal proofs from arbitrary proofs. Based on this definition, any path in a normal proof starts with an axiom/lexicon rule, then passes through a (possibly empty) sequence of elimination rules as the major premiss, followed by a (possibly empty) sequence of introduction rules, ending either in the minor premiss of an elimination rule or in the conclusion of the proof.

To demonstrate HTLG is normalising, we define a set of conversion rules --- functions from derivations $D$ to derivations $D'$ --- such that repeated application of the rules terminates in a normal derivation. 

%



Figure \ref{fig:conv} shows the conversion rules.\footnote{The rule for the $\slash$ directly parallels that for $\backslash$, modulo directionality.} Note that, given the condition on the elimination rules, $N$ and $M$ cannot share free variables and that Lemma~\ref{lem:subst} therefore guarantees the reductions transform proofs into proofs.

There is an additional complication for the $\himpl$ case: we replace a $\beta$ redex $((\lambda x. M)\, N)$ by its contractum $M[x:= N]$. We therefore need to verify that we can construct $\delta_3'$ from $\delta_3$. However, inspection of the rules shows as that all rules in $\delta_3$ can be performed in $\delta_3'$, with the possible exception of a $\beta$ reduction on the redex $((\lambda x. M)\, N)$ (or more precisly its trace $(\lambda x. M')\, N'$ obtained by reducing $M$ to $M'$ and $N$ to $N'$ in any number of steps). However, in this case we can simply remove the $\beta$ reduction from $\delta_3'$ to obtain a valid proof. Finally, the rules in $\delta_3'$ are therefore those in $\delta_3$ with possibly a single $\beta$ reduction removed and with $P$ either identical to $P'$ or $\beta$-reducible to it in one step.\footnote{In case $((\lambda x. M)\, N)$ is an $\eta$ redex and we perform an $\eta$ reduction on it, its trace must be of the form $((\lambda x. (M'\, x))\, N')$, and the $\eta$ redex of this term is identical to its $\beta$ redex.}
\moot{I'm not really happy with this last paragraph, but we need something like: performing beta reduction to a newly formed beta redex preserves provability.}





In what follows, we assume that $\beta$ reduction applies on an as-needed basis, ignoring its application for simplicity of presentation. 

\begin{figure}
\begin{center}
	\begin{tabular}{ccc}
	\infer*[\delta_3]{P:C}{\infer[\backslash E]{(N+M)^{\str}:B}{\infer*[\delta_1]{N^{\str}:A}{} & \infer[\backslash I]{M^{\str}:A\backslash B}{\infer*[\delta_2]{(x+M)^{\str}:B}{x^{\str}:A}}}} & $\;\;\;\rightsquigarrow\;\;\;$ &
	\infer*[\delta_3]{P:C}{\infer*[\delta_2\bo x:=N\bc]{(N+M)^{\str}:B}{\infer*[\delta_1]{N^{\str}:A}{}}} \\
	\\
	\infer*[\delta_3]{P:C}{\infer[\multimap E]{((\lambda x. M)N)^{\beta}:B}{\infer*[\delta_1]{N^{\alpha}:A}{} & \infer[\multimap I]{(\lambda x. M)^{\alpha\rightarrow\beta}:A\multimap B}{\infer*[\delta_2]{M^{\beta}:B}{x^{\alpha}:A}}}} & $\;\;\;\rightsquigarrow\;\;\;$ &
  \infer*[\delta_3']{P':C}{\infer*[\delta_2\bo x^{\alpha}:=N^{\alpha}\bc]{(M[x:=N])^{\beta}:B}{\infer*[\delta_1]{N^{\alpha}:A}{}}}
	\end{tabular}
\end{center}
%
	
	
	\caption{Conversion Rules}
	\label{fig:conv}
\end{figure}

\begin{theorem}\label{thm:norm}
	HTLG is strongly normalising.
\end{theorem}

\paragraph{Proof} To show strong normalisation, we need to show that there are no infinite reduction sequences.
Since each reduction reduces the size of the proof, this is trivial. \qedsym

\subsection{Properties and consequences of normalisation}

 \begin{theorem}\label{thm:confl}
	Normalisation for HTLG proofs is confluent.
\end{theorem}

\paragraph{Proof} It is easy to show weak confluence: whenever a proof can be reduced by two different reductions $R_1$ and $R_2$, then reducing either redex will preserve the other redex, and $R_1$ followed by $R_2$ will produce the same proof as $R_2$ followed by $R_1$. By Newman's Lemma \cite{newman42theories}, we know that a rewrite system which is strongly normalising and weakly confluent is strongly confluent. Theorem~\ref{thm:norm} and weak confluence therefore entail strong confluence. \qedsym

\begin{corollary} HTLG proofs have a unique normal form.
\end{corollary}

\paragraph{Proof} Immediate by Theorem~\ref{thm:norm} and Theorem~\ref{thm:confl}. Note that uniqueness is up to beta-eta equivalence or, alternatively, each HTLG proof has a unique normal form proof with a term in beta-eta normal form\footnote{As is usual in the lambda calculus, we do not distinguish alpha-equivalent lambda terms.}. \qedsym

\begin{corollary} HTLG satisfies the subformula property.
	
\end{corollary}

\paragraph{Proof} The only case requiring special attention is the \textit{Lex} rule.
\[
\infer[\textit{Lex}]{p^{\str}:w \vdash M:A}{}
\]
In the \textit{Lex} rule the antecedent formula $p^{\str}:w$, with $p$ a variable unique in the proof corresponding to this rule, the atomic formula $w$ corresponds to the $A$ formula assigned by the lexicon. For determining subformulas, we therefore treat each occurrence of $p^{\str}:w$ as the formula $A$ assigned to it by the lexicon.

The subformula is then a direct consequence of normalisation (Theorem~\ref{thm:norm}). In a normal form proof, every formula is either a subformula of one of the hypotheses or a subformula of the conclusion. \qedsym


\begin{corollary} HTLG is a conservative extension of the (product-free) Lambek calculus. 
\end{corollary}

\begin{corollary} HTLG is a conservative extension of lambda grammars.
\end{corollary}

\paragraph{Proof} These are simple corollaries of the subformula property. All theorems of the Lambek calculus are theorems of HTLG and HTLG restricted to Lambek calculus formulas proves only Lambek calculus theorems. The same holds for lambda grammars. \qedsym








Given that we only consider linear lambda terms, HTLG proofs have a number of beta reductions bounded from above by the total number of abstractions in the proof (those in the lexical leaves plus those in the introduction rules). Therefore, decidability follows from the subformula property. However, we will give a more detailed complexity analysis in Section~\ref{sec:compl}.

A drawback of natural deduction is that normalisation becomes more complicated when we want to add other logical connectives, such as the Lambek calculus product `$\bullet$'. Instead of solving these problems in natural deduction, we introduce two alternative calculi for HTLG: a sequent calculus and  a proof net calculus.

\section{Sequent calculus}
\label{sec:seq}

HTLG is most easily presented in the form of a natural deduction calculus.
In this section, we provide a term-labeled sequent calculus for it as well.

\begin{figure}
\begin{center}
\begin{tabular}{ccc}
  \infer[\textit{Cut}]{\Gamma,\Delta\seq N[M]:C}{\Delta\seq M^{\alpha}:A & \Gamma,x^{\alpha}:A\seq N[x]:C}  &\;\;  & \infer[\textit{Ax}]{x:A \seq x:A}{} \\[2.5mm]
  \infer[/L]{\Gamma,\Delta,p^{\str}:B/A \seq N[p+M]:C}{\Delta\seq M^{\str}:A & \Gamma,q^{\str}:B\seq N[q]:C} & &
  \infer[/R]{\Gamma \seq M^{\str}:B/A}{\Gamma, p^{\str}:A \seq (M+p)^{\str}:B} \\[2.5mm]
  \infer[\backslash L]{\Gamma,\Delta,p^{\str}:A\backslash B \seq N[M+p]:C}{\Delta\seq M^{\str}:A & \Gamma,q^{\str}:B\seq N[q]:C} & &
  \infer[\backslash R]{\Gamma \seq M^{\str}:A\backslash B}{\Gamma, p^{\str}:A \seq (p+M)^{\str}:B} \\[2.5mm]
  \infer[\himpl L]{\Gamma,\Delta,x^{\alpha\rightarrow\beta}:A\himpl B \seq N[(x \,M)]:C}{\Delta\seq M^{\alpha}:A & \Gamma,y^{\beta}:B\seq N[y]:C} & &
  \infer[\himpl R]{\Gamma \seq (\lambda x. M)^{\alpha\rightarrow\beta}:A\himpl B}{\Gamma, x^{\alpha}:A \seq M^{\beta}:B} \\[2.5mm]
  \infer[\textit{Lex}]{\Gamma,p^{\str}:w\seq M[N[p]^{\alpha}]:C}{\Gamma,x^{\alpha}:A \seq M[x]:C} &&
  \infer[\beta\eta]{\Gamma \seq M:C}{\Gamma \seq M':C} 
\end{tabular}
\end{center}
\caption{Sequent calculus rules for HTLG}
\label{fig:seq}
\end{figure}

Figure~\ref{fig:seq} shows the sequent calculus rules for HTLG. The rules for $\multimap$ are a standard way of adding lambda term labeling to sequent calculus proofs.

The lexicon rule \textit{Lex}, when read from premisses to conclusion in a forward chaining proof search strategy, replaces a variable $x$ of type $\alpha$ by a term $N[z]$ of the same type, where $z$ is the only free variable in $N$ and $N[z]:A$ is assigned by the lexicon to the word corresponding to $p$. Proof-theoretically, the sequent calculus \textit{Lex} is just a combination of the natural deduction \textit{Lex} rule with the \textit{Cut} rule. 

The $\beta\eta$ rule applies when $M\equiv_{\beta\eta} M'$ in the lambda calculus. Since we are only interested in proof terms modulo $\beta\eta$ equivalence, we generally apply the $\beta\eta$ rule only when $M$ is the $\beta\eta$ normal form of $M$, or, alternatively, when $M$ is obtained from $M'$ by a single beta reduction, as we did with the $\beta\eta$ rule of natural deduction.
We will also assume that whenever a rule produces a term containing a $\beta$ redex (only the \textit{Cut} and \textit{Lex} rules can produce $\beta$ redexes\footnote{The $\beta\eta$ rule cannot create redexes when we assume its conclusion term $M$ must be in long normal form.}), we immediately transform it into $\beta$ normal form using the $\beta\eta$ rule. 



\subsection{Equivalence with natural deduction}

\begin{lemma} $\Gamma \vdash M:C$ iff and only if $\Gamma \seq M:C$.
\end{lemma}
  
\paragraph{Proof}
All rules in the right hand column of Figure~\ref{fig:seq} correspond directly to a rule in the natural deduction calculus of Figure~\ref{fig:nd} (the right rules for the connectives correspond to the introduction rules, and the $\textit{Ax}$ and $\beta\eta$ rules are the same in both calculi). For the remaining rules, we simply show that they are derived rules in the other calculus.

[Sequent calculus to natural deduction]
For the translation of sequent calculus proofs to natural deduction proofs, it therefore suffices to translate the $\textit{Cut}$, $/\textit{L}$, $\backslash\textit{L}$, $\multimap\textit{L}$ and $\textit{Lex}$ rules. 

[$\textit{Cut}$] The $\textit{Cut}$ rule in sequent calculus corresponds to the application of the Substitution Lemma (Lemma~\ref{lem:subst}) in natural deduction.

[$/\textit{L}$] Given a natural deduction proof $\delta_1$ of $\Delta \vdash M^{\str}:A$ and a natural deduction proof $\delta_2$ of $\Gamma, q^{\str}:B \seq N[q]:C$, we can produce a natural deduction proof of $\Gamma,\Delta,p^{\str}:B/A\vdash N[p+M]:C$ as follows (where $\textit{SL}$ is the Substitution Lemma).

\[
\infer[\textit{SL}]{\Gamma,\Delta,p^{\str}:B/A\nd N[p+M]:C}{
 \infer[/\textit{E}]{\Delta,p^{\str}:B/A \nd (p+M)^{\str}:B}{
  \infer[\textit{Ax}]{p^{\str}:B/A\nd p^{\str}:B/A}{}
  & \infer*[\delta_1]{\Delta\nd M^{\str}:A}{}}
& \infer*[\delta_2]{\Gamma, q^{\str}:B \nd N[q]:C}{}}
\]

[$\backslash\textit{L}$] Symmetric.

[$\multimap\textit{L}$]  Similar to the $/\textit{L}$ case, we can derive instantiations of the $\himpl\textit{L}$ rule using a combination of an $A\himpl B$ axiom, the $\himpl\textit{E}$ rule, and the Substitution Lemma (represented by $\textit{SL}$ in the proof below).

\[
\infer[\textit{SL}]{\Gamma,\Delta,x^{\alpha\rightarrow \beta}:A\himpl B\nd N[(x\, M)]:C}{
 \infer[\himpl\textit{E}]{\Delta,x^{\alpha\rightarrow\beta}:A\himpl B \nd (x\, M)^{\beta}:B}{
  \infer[\textit{Ax}]{x^{\alpha\rightarrow\beta}:A\himpl B \nd x^{\alpha\rightarrow\beta}:A\himpl B}{}
  & \infer*[\delta_1]{\Delta\nd M^{\alpha}:A}{}}
& \infer*[\delta_2]{\Gamma, y^{\beta}:B \nd N[y]:C}{}}
\]

[$\textit{Lex}$] For the $\textit{Lex}$, we need to show that we can transform a natural deduction proof $\delta$ of $\Gamma,x:A \seq M[x]:C$ into a proof of $\Gamma, p:w \seq M[N[p]]:A$. This is done using the substitution lemma for $\delta$ and the corresponding natural deduction $\textit{Lex}$ rule $p:w \seq N[p]:A$.

[Natural deduction to sequent calculus] 
For the translation of natural deduction proofs to sequent calculus proofs, it suffices to translate the $/\textit{E}$, $\backslash\textit{E}$, $\multimap\textit{E}$ and $\textit{Lex}$ rules.

[$/\textit{E}$]  We translate the $/\textit{E}$ rule

\[
\infer[/\textit{E}]{\Gamma,\Delta\nd (M+N)^{\str}:A}{\infer*[\delta_1]{\Gamma\nd M^{\str}:A/B}{} & \infer*[\delta_2]{\Delta\nd N^{\str}:B}{}}
\]

as follows.

\[
\infer[\textit{Cut}]{\Gamma,\Delta \seq (M+N)^{\str}:A}{
  \infer*[\delta'_1]{\Gamma\seq M^{\str}:A/B}{}
&  \infer[/\textit{L}]{\Delta,x^{\str}:A/B \seq (x+N)^{\str}:A}{\infer*[\delta'_2]{\Delta\seq N^{\str}:B}{} & \infer[\textit{Ax}]{z^{\str}:A\seq z^{\str}:A}{}}}
\]

[$\backslash \textit{E}$] Symmetric.

[$\himpl\textit{E}$]  We translate the $\himpl\textit{E}$ rule

\[
\infer[\himpl\textit{E}]{\Gamma,\Delta\nd (M\,N)^{\alpha}:A}{\infer*[\delta_1]{\Gamma\nd M^{\beta\rightarrow\alpha}:B\himpl A}{} & \infer*[\delta_2]{\Delta\nd N^{\beta}:B}{}}
\]

as follows.

\[
\infer[\textit{Cut}]{\Gamma,\Delta \seq (M\, N)^{\alpha}:A}{
  \infer*[\delta'_1]{\Gamma\seq M^{\beta\rightarrow\alpha}:B\himpl A}{}
  &  \infer[\himpl\textit{L}]{\Delta,x^{\beta\rightarrow\alpha}:B\himpl A \seq (x\, N)^{\alpha}:A}{\infer*[\delta'_2]{\Delta\seq N^{\beta}:B}{} & \infer[\textit{Ax}]{z^{\alpha}:A\seq z^{\alpha}:A}{}}}
\]
[\textit{Lex}] We translate the $\textit{Lex}$ rule

\[
\infer[\textit{Lex}]{p^{\str}:w\vdash M^{\alpha}:A}{}
\]

\noindent as follows.

\[
\infer[\textit{Lex}]{p^{\str}:w\seq M:A}{\infer[\textit{Ax}]{x^{\alpha}:A \seq x^{\alpha}:A}{}}
\]

\subsection{Cut elimination}
\label{sec:seqcutelim}

We show that the sequent calculus for HTLG presented in Figure~\ref{fig:seq} satisfies cut elimination. That is, any HTLG sequent which can be derived using the \textit{Cut} rule can also be derived without it.

The proof is fairly standard and follows the proof of \citet{lambek} for the Lambek calculus. A \textit{Cut} rule looks as follows.

\[
\infer[\textit{Cut}]{\Gamma,\Delta\seq N[M]:C}{\infer[R_l]{\Gamma\seq M:A}{\infer*[\delta_1]{}{}} & \infer[R_r]{\Delta,x:A\seq N[x]:C}{\infer*[\delta_2]{}{}}}
\]

As is usual with cut elimination proofs, there are many commutative cases, where at least one of $R_l$ and $R_r$ does not create the cut formula $A$ (these are treated in cases \emph{b} and \emph{c} below). We assume that neither $\delta_1$ nor $\delta_2$ contains any cut rules and proceed by induction on the lexicographic ordering $\langle \textit{degree},\textit{depth}\rangle$, where \emph{degree} is the number of connectives in $A$ and \emph{depth} is the sum of the depth in $\delta_1$ and $\delta_2$ of the rules introducing the cut formula $A$ in the subproofs. This is either the right rule for the main connective of $A$ or an axiom in $R_l$ and either the left rule for the main connective of $A$ or an axiom in $R_r$. It is important that the \emph{Lex} rule can introduce $A$ in neither in $R_l$ (since the premiss of the \emph{Lex} rule has the same formula $A$, the conclusion of the rule can never be the topmost occurrence of $A$ in the proof) nor in $R_r$ (since the formula $w$ cannot appear on the right hand side of the sequent symbol). We therefore show that we can either eliminate the cut immediately (case a),  replace it by a cut on its subformulas (case d, where $\textit{depth}=0$ and we reduce the degree of the new cuts), or move the cut formula up, reducing the \textit{depth} (case b and c, depending on whether we move the cut rule up with respect to $R_l$ or $R_r$).


a. The base cases are simple. When $R_l$ is an axiom, we have $\Delta = y:A$ and $M=y$ and the conclusion of the \emph{Cut} rule is $\Gamma,y:A \seq N[y]:C$. We can remove the \textit{Cut} and obtain a proof of the same term (up to alpha equivalence) as follows.

\[
\infer[\textit{Cut}]{\Gamma,y:A \seq N[y]:C}{\infer[\textit{Ax}]{y:A \seq y:A}{} & \infer*[\delta_2]{\Gamma,x:A\seq N[x]:C}{}} \qquad \leadsto \qquad
 \infer*[\delta_2]{\Gamma,x:A\seq N[x]:C}{}
\]

 Similarly, when $R_r$ is an axiom, $N[]$ is the empty context term, $\Gamma$ is the empty formula sequence and $C=A$, making the conclusion of the \emph{Cut} rule $\Delta \seq M:A$ which is the same as the conclusion of $R_l$, so we can remove the \emph{Cut} by taking the proof $\delta_1$.

\[
\infer[\textit{Cut}]{\Delta\seq M:A}{\infer*[\delta_1]{\Delta\seq M:A}{} & \infer[\textit{Ax}]{x:A\seq x:A}{}}
\qquad \leadsto \qquad
\infer*[\delta_1]{\Delta\seq M:A}{}
\]

b. \emph{$R_l$ does not create the cut formula $M:A$.}

If $R_l$ does not create the cut formula, then $R_l$ must be $\backslash\textit{L}$, $/\textit{L}$, $\himpl\textit{L}$, $\beta\eta$ or \textit{Lex}. We show that in each case, we can move the application of the cut rule up with respect to $R_l$ while keeping the depth with respect to $R_r$ the same, thereby reducing the \emph{depth} parameter of the induction. We look at all the cases.

-- [$\backslash\textit{L}$] The case for $\backslash\textit{L}$ looks as follows.

\[
\infer[\textit{Cut}]{\Gamma,\Delta,\Delta',q:B\backslash D\seq N[M[P+q]]:A}{
  \infer[\backslash\textit{L}]{\Gamma,\Delta',q:B\backslash D\seq M[P+q]:A}{
    \infer*[\delta_1]{\Delta'\seq P:B}{}
  & \infer*[\delta_2]{\Gamma,r:D\seq M[r]:A}{}
  }
 & \infer*[\delta_3]{\Delta,x:A\seq N[x]:C}{}}
\]

We can move the cut rule up as follows.

\[
\infer[\backslash\textit{L}]{\Gamma,\Delta,\Delta',q:B\backslash D\seq N[M[P+q]]:A}{
  \infer*[\delta_1]{\Delta'\seq P:B}{}
  & \infer[\textit{Cut}]{\Gamma,\Delta,r:D \seq N[M[r]]:C}{
       \infer*[\delta_2]{\Gamma,r:D\seq M[r]:A}{}
    &  \infer*[\delta_3]{\Delta,x:A\seq N[x]:C}{}
    }
}
\]

We need to be careful with the variable names, ensuring $r$ doesn't occur in $\Delta$ and $N$ for the proof above to be well-formed. We will simply assume here and elsewhere that all variables assigned to formulas are unique in the proof. This can easily be guaranteed by renaming.

-- [$/\textit{L}$] The case for $/\textit{L}$ is symmetric to the case for $\backslash\textit{L}$.

-- [$\himpl\textit{L}$] The case for $\himpl\textit{L}$ looks as follows.

\[
\infer[\textit{Cut}]{\Gamma,\Delta,\Delta',y:B\himpl D\seq N[M[(y\, P)]]:A}{
  \infer[\himpl\textit{L}]{\Gamma,\Delta',y:B\himpl D\seq M[(y\, P)]:A}{
    \infer*[\delta_1]{\Delta'\seq P:B}{}
  & \infer*[\delta_2]{\Gamma,z:D\seq M[z]:A}{}
  }
 & \infer*[\delta_3]{\Delta,x:A\seq N[x]:C}{}}
\]

We can move the cut up as follows.

\[
\infer[\himpl\textit{L}]{\Gamma,\Delta,\Delta',y:B\himpl D\seq N[M[(y\, P)]]:A}{
  \infer*[\delta_1]{\Delta'\seq P:B}{}
  & \infer[\textit{Cut}]{\Gamma,\Delta, z:D \seq N[M[z]]:C}{
       \infer*[\delta_2]{\Gamma,z:D\seq M[z]:A}{}
    &  \infer*[\delta_3]{\Delta,x:A\seq N[x]:C}{}
    }
}
\]

-- [$\textit{Lex}$]
The case for \textit{Lex} looks as follows.

\[
\infer[\textit{Cut}]{\Gamma,\Delta,z:w\seq P[M[N[z]]]:C}{
    \infer[\textit{Lex}]{\Gamma,z:w \seq M[N[z]]:A}{\infer*[\delta_1]{\Gamma,y:B\seq M[y]:A}{}} &
      \infer*[\delta_2]{\Delta,x:A\seq P[x]:C}{}
        }
\]

We can move the cut up in the following way.
\[
\infer[\textit{Lex}]{\Gamma,\Delta,z:w\seq P[M[N[z]]]:C}{
   \infer[\textit{Cut}]{\Gamma,\Delta,y:B\seq P[M[y]]:C}{
     \infer*[\delta_1]{\Gamma,y:B\seq M[y]:A}{} 
  &  \infer*[\delta_2]{\Delta,x:A\seq P[x]:C}{}
   }
}
\]

-- [$\beta\eta$]
The case for $\beta\eta$ looks as follows.

\[
\infer[\textit{Cut}]{\Gamma,\Delta\seq P[M[N]]:C}{
    \infer[\beta\eta]{\Gamma \seq M[N]:A}{\infer*[\delta_1]{\Gamma \seq M[N']:A}{}} &
      \infer*[\delta_2]{\Delta,x:A\seq P[x]:C}{}
        }
\]

We can move the cut up in the following way.

\[
\infer[\beta\eta]{\Gamma,\Delta \seq P[M[N]]:A}{
\infer[\textit{Cut}]{\Gamma,\Delta\seq P[M[N']]:C}{
    \infer*[\delta_1]{\Gamma \seq M[N']:A}{}
  & \infer*[\delta_2]{\Delta,x:A\seq P[x]:C}{}
        }}
\]

c. \emph{$R_r$ does not create the cut formula $x:A$.}

There are many cases to consider here: rule $R_r$ can be one of the right rules, one of the left rules for a formula other than $A$, and it can also be \textit{Lex} or $\beta\eta$.

-- [$\backslash\textit{R}$] The $\backslash\textit{R}$ case looks as follows.

\[
\infer[\textit{Cut}]{\Gamma,\Delta\seq N[M]:B\backslash C}{
  \infer*[\delta_1]{\Gamma\seq M:A}{}
  & \infer[\backslash\textit{R}]{\Delta,x:A\seq N[x]:B\backslash C}{
       \infer*[\delta_2]{\Delta,q:B,x:A\seq q+N[x]:C}{}
    }
}
\]
We can move the cut up as follows.
\[
\infer[\backslash\textit{R}]{\Gamma,\Delta \seq N[M]:B\backslash C}{
  \infer[\textit{Cut}]{\Gamma,\Delta, q:B\seq q+N[M]:C}{
     \infer*[\delta_1]{\Gamma\seq M:A}{}
   & \infer*[\delta_2]{\Delta,q:B,x:A\seq q+N[x]:C}{}
    }
}
\]

-- [$/\textit{R}$] The case for $/\textit{R}$ is symmetric.

-- [$\himpl\textit{R}$] The $\himpl\textit{R}$ case looks as follows.

\[
\infer[\textit{Cut}]{\Gamma,\Delta\seq \lambda y. N[M]:B\himpl C}{
  \infer*[\delta_1]{\Gamma\seq M:A}{}
  & \infer[\himpl\textit{R}]{\Delta,x:A\seq \lambda y. N[x]:B\himpl C}{
       \infer*[\delta_2]{\Delta,y:B,x:A\seq N[x]:C}{}
    }
}
\]
We can move the cut up as follows.
\[
\infer[\himpl\textit{R}]{\Gamma,\Delta \seq \lambda y. N[M]:B\himpl C}{
  \infer[\textit{Cut}]{\Gamma,\Delta, y:B\seq N[M]:C}{
     \infer*[\delta_1]{\Gamma\seq M:A}{}
   & \infer*[\delta_2]{\Delta,y:B,x:A\seq N[x]:C}{}
    }
}
\]
-- [$\textit{Lex}$] If the last rule in $R_r$ is a \textit{Lex} rule, we are in the following situation.
\[
\infer[\textit{Cut}]{\Gamma,\Delta,z:w\seq M[P][N[z]]:C}{
  \infer*[\delta_1]{\Delta\seq P:A}{}
  & \infer[\textit{Lex}]{\Gamma,x:A,z:w\seq M[x][N[z]]:C}{
    \infer*[\delta_2]{\Gamma,x:A,y:B \seq M[x][y]:C}{}
  }
  }
\]
We can move the cut rule up as follows. This is a valid proof because $z$ is the only free variable in $N$.
\[
\infer[\textit{Lex}]{\Gamma,\Delta,z:w\seq M[P][N[z]]:C}{
\infer[\textit{Cut}]{\Gamma,\Delta,y:B\seq M[P][y]:C}{
  \infer*[\delta_1]{\Delta\seq P:A}{}
& \infer*[\delta_2]{\Gamma,x:A,y:B \seq M[x][y]:C}{}
}}
\]

-- [$\beta\eta$] If the last rule in $R_r$ is a $\beta\eta$ rule, we are in the following situation.
\[
\infer[\textit{Cut}]{\Gamma,\Delta\seq M[P]:C}{
  \infer*[\delta_1]{\Delta\seq P:A}{}
  & \infer[\beta\eta]{\Gamma, x:A\seq M[x]:C}{
    \infer*[\delta_2]{\Gamma,x:A \seq M'[x]:C}{}
  }
  }
\]
We can move the cut rule up as follows. This is a valid proof because $M$ and $M'$ have the same free variables.
\[
\infer[\beta\eta]{\Gamma,\Delta\seq M[P]:C}{
\infer[\textit{Cut}]{\Gamma,\Delta\seq M'[P]:C}{
  \infer*[\delta_1]{\Delta\seq P:A}{}
& \infer*[\delta_2]{\Gamma,x:A \seq M'[x]:C}{}
}}
\]
-- [$\backslash\textit{L}$] If the last rule is $\backslash\textit{L}$ we are in the following situation.
\[
\infer[\textit{Cut}]{\Gamma,\Delta,\Delta',p:B\backslash D \seq M[P][Q+p]:C}{
  \infer*[\delta_1]{\Delta\seq P:A}{}
  &\infer[\backslash\textit{L}]{\Gamma,\Delta',x:A,p:B\backslash D\seq M[x][Q+p]:C}{
    \infer*[\delta_2]{\Delta'\seq Q:B}{}
  & \infer*[\delta_3]{\Gamma,x:A,q:D\seq M[x][q]:C}{}
  }
}
\]
We can transform the proof as follows.
\[
\infer[\backslash\textit{L}]{\Gamma,\Delta,\Delta',p:B\backslash D \seq M[P][Q+p]:C}{
  \infer*[\delta_2]{\Delta'\seq Q:B}{}
  & \infer[\textit{Cut}]{\Gamma,\Delta,q:D\seq M[P][q]:C}{
    \infer*[\delta_1]{\Delta\seq P:A}{}
  & \infer*[\delta_3]{\Gamma,x:A,q:D\seq M[x][q]:C}{}
    }
}
\]

-- [$/\textit{L}$] The case for $/\textit{L}$ is symmetric.

-- [$\himpl\textit{L}$] If the last rule is $\himpl\textit{L}$ we are in the following situation.
\[
\infer[\textit{Cut}]{\Gamma,\Delta,\Delta',y:B\himpl D \seq M[P][(y\, N)]:C}{
  \infer*[\delta_1]{\Delta\seq P:A}{}
  &\infer[\himpl\textit{L}]{\Gamma,\Delta',x:A,y:B\himpl D\seq M[x][(y\, N)]:C}{
    \infer*[\delta_2]{\Delta'\seq N:B}{}
  & \infer*[\delta_3]{\Gamma,x:A,z:D\seq M[x][z]:C}{}
  }
}
\]
We can transform the proof as follows.
\[
\infer[\backslash\textit{L}]{\Gamma,\Delta,\Delta',y:B\backslash D \seq M[P][(y\, N)]:C}{
  \infer*[\delta_2]{\Delta'\seq N:B}{}
  & \infer[\textit{Cut}]{\Gamma,\Delta,z:D\seq M[P][z]:C}{
    \infer*[\delta_1]{\Delta\seq P:A}{}
  & \infer*[\delta_3]{\Gamma,x:A,z:D\seq M[x][z]:C}{}
    }
}
\]

d. \emph{Both $R_l$ and $R_r$ create the cut formula $A$.}

Only a combination of a left rule and a right rule for the same connective can create a cut formula. So we only have the case for $\backslash$, $/$ and $\himpl$ to consider.

-- [$\backslash$] The case for $\backslash$ looks as follows.

\[
\infer[\textit{Cut}]{\Gamma,\Delta,\Gamma'\seq N[P+M]:C}{
  \infer[\backslash\textit{R}]{\Gamma\seq M:A\backslash B}{\infer*[\delta_1]{\Gamma,p:A\seq p+M:B}{}} &
  \infer[\backslash\textit{L}]{\Gamma',\Delta,q:A\backslash B \seq N[P+q]:C}{
    \infer*[\delta_2]{\Delta \seq P:A}{}
  & \infer*[\delta_3]{\Gamma',r:B \seq N[r]:C}{}
  }
}
\]

We can replace the cut on $A\backslash B$ by two cuts of strictly lower degree on the subformulas $A$ and $B$ as follows.

\[
\infer[\textit{Cut}]{\Gamma,\Delta,\Gamma'\seq N[P+M]:C}{
  \infer[\textit{Cut}]{\Gamma,\Delta\seq P+M:B}{
    \infer*[\delta_2]{\Delta \seq P:A}{}
  & \infer*[\delta_1]{\Gamma,p:A\seq p+M:B}{}
  }
  & \infer*[\delta_3]{\Gamma',r:B \seq N[r]:C}{}
}
\]

-- [$/$] The case for $/$ is symmetric to the case for $\backslash$

-- [$\himpl$] 
Finally, the case for $\himpl$ looks as follows.
\[
\infer[\textit{Cut}]{\Gamma,\Delta,\Gamma'\seq N[((\lambda x.M[x])\, P)]:C}{
  \infer[\himpl\textit{R}]{\Gamma\seq \lambda x. M[x]:A\himpl B}{\infer*[\delta_1]{\Gamma,x:A\seq M[x]:B}{}} &
  \infer[\himpl\textit{L}]{\Gamma',\Delta,y:A\himpl B \seq N[(y\, P)]:C}{
    \infer*[\delta_2]{\Delta \seq P:A}{}
  & \infer*[\delta_3]{\Gamma',z:B \seq N[z]:C}{}
  }
}
\]


We show that we can replace the cut by two cuts on the immediate subformulas as follows producing the required redex $N[M[P]]$ as follows.
\[
\infer[\textit{Cut}]{\Gamma,\Delta,\Gamma'\seq N[M[P]]:C}{
  \infer[\textit{Cut}]{\Gamma,\Delta\seq M[P]:B}{
    \infer*[\delta_2]{\Delta \seq P:A}{}
  & \infer*[\delta_1]{\Gamma,x:A\seq M[x]:B}{}
    }
  & \infer*[\delta_3]{\Gamma',z:B \seq N[z]:C}{}
}
\]

We have removed the Cut redex of the previous proof and replaced it with its contractum. We need to verify that this preserves the proof rules occurring below the two reduced new cut rule applications. The only rules potentially affected are $\beta\eta$ rules reducing $\lambda x$ for the given occurrence of the $\lambda$ abstraction; all other rules for operate on one of the terms $N$, $M$, $P$ of  $N[(\lambda x. M)\, P]$ --- $N$ can, depending on its type, be applied or concatenated to another term, or have one of its rightmost or leftmost string variables removed, whereas all three of $M$, $N$, and $P$ can have lexical substitution or $\beta\eta$ reductions take place inside them, producing a term $N'[(\lambda x. M')\, P']$. For linear lambda terms, only the $\beta$ and $\eta$ rule can remove an abstraction. However, these rules replace $(\lambda x. M'[x])\, P'$ by $M'[P']$ (for the $\beta$ case), and $(\lambda x. (M' x))\, P'$ by $(M' P')$ (for the $\eta$ case). In either case, we can simply remove this $\beta$ or $\eta$ reduction from the proof since it has become superfluous.  \qed
\section{Proof nets}
\label{sec:pn}



Proof nets are a graph theoretic representation of proofs introduced for linear logic by \citet{Girard}. Proof nets remove the possibility of `boring' rule permutations as they occur in the sequent calculus or natural deduction,\footnote{For natural deduction, rule permutations are a problem only for the $\bullet E$ and the $\Diamond E$ rules.} solving the so-called problem of `spurious ambiguity' in type-logical grammars. 

We generally define proof nets as part of a larger class called \emph{proof structures}. Proof nets are those proof structures which correspond to sequent (or natural deduction) proofs. We can distinguish proof nets from other proof structures by means of a correctness condition. As a guiding intuition, we have the following correspondence between sequent calculus/natural deduction proofs and proof nets.

\medskip
\begin{tabular}{rl}
logical rule & = link + correctness condition \\
proof 	(proof net) & = proof structure + correctness condition
\end{tabular}
\medskip

Another way to see this is that proof structures are \emph{locally} correct whereas proof nets are \emph{globally} correct as well.

A more procedural interpretation of this is that a proof structures represent the search space for proofs. 

\subsection{Proof structures}

\begin{definition} A \emph{link} is tuple consisting of a \emph{type} (tensor or par), an \emph{index} (from a fixed alphabet $I$, indicating the family of connectives it belongs to), a \emph{list of premisses}, a \emph{list of conclusions}, and an optional \emph{main node} (either one of the conclusions or one of the premisses).
\end{definition}

A link is essentially a labelled hyperedge connecting a number of vertices in a hypergraph. The premisses of a link are drawn left-to-right above the central node,
whereas the conclusions are drawn left-to-right below the central node. A par link displays the central node as a filled circle, whereas a tensor uses an open circle. For hybrid type-logical grammars, the set of indices is $\{ \epsilon, +, @, \lambda \}$. The constructor $\epsilon$ represents the empty string (it doesn't correspond to a logical connective, although we can add one if desired). The
(non-associative) Lambek calculus implications ($\ldl$, $\ldr$) use the term constructor `$+$' for their links (in a multimodal context we can have multiple instances of `$+$', for example, `$+_1$', `$+_2$', but this doesn't change much), whereas the lambda grammar implication ($\himpl$) uses links labeled with $@$ (representing application, for its tensor link) and $\lambda$ (representing abstraction, for its par link). 

\begin{table}
\begin{center}
\begin{tikzpicture}[scale=0.75]
\node (labl) at (3em,7.5em) {$[\ldr E]$};
\node (ab) at (3em,9.8em) {$C$};
\node (a) at (0,14.6em) {$C\ldr B$};
\node (aa) at (0.7em,14.2em) {};
\node (b) at (6em,14.75em) {$B$};
\node[tns] (c) at (3em,12.668em) {};
\node (plus) at (3em,12.668em) {$+$};
\draw (c) -- (ab);
\draw (c) -- (aa);
\draw (c) -- (b);
\node (labl) at (3em,-2.0em) {$[\ldr I]$};
\node (pa) at (0,0) {$C\ldr B$};
\node (pat) at (0.7em,0.44em) {};
\node[par] (pc) at (3em,1.732em) {};
\node at (3em,1.732em) {\nodeindex{+}};
\node (pb) at (6em,0.15em) {$B$};
\node (pd) at (3em,4.8em) {$C$};
\draw (pc) -- (pb);
\draw (pc) -- (pd);
\path[>=latex,->]  (pc) edge (pat);
\node (labl) at (13em,7.5em) {$[\ldl E]$};
\node (ab) at (13em,9.8em) {$C$};
\node (a) at (16em,14.6em) {$A\ldl C$};
\node (aa) at (15.3em,14.2em) {};
\node (b) at (10em,14.75em) {$A$};
\node[tns] (c) at (13em,12.668em) {};
\node (plus) at (13em,12.668em) {$+$};
\draw (c) -- (ab);
\draw (c) -- (aa);
\draw (c) -- (b);
\node (labl) at (13em,-2.0em) {$[\ldl I]$};
\node (pa) at (16em,0) {$A\ldl C$};
\node (pat) at (15.3em,0.44em) {};
\node[par] (pc) at (13em,1.732em) {};
\node at (13em,1.732em) {\nodeindex{+}};
\node (pb) at (10em,0.15em) {$A$};
\node (pd) at (13em,4.8em) {$C$};
\draw (pc) -- (pb);
\draw (pc) -- (pd);
\path[>=latex,->]  (pc) edge (pat);
%
\node (labl) at (23em,7.5em) {$[\himpl E]$};
\node (ab) at (23em,9.8em) {$C$};
\node (a) at (20em,14.6em) {$B\himpl C$};
\node (aa) at (20.7em,14.2em) {};
\node (b) at (26em,14.75em) {$B$};
\node[tns] (c) at (23em,12.668em) {};
\node (appl) at (23em,12.668em) {$@$};
\draw (c) -- (ab);
\draw (c) -- (aa);
\draw (c) -- (b);
\node (labl) at (23em,-2.0em) {$[\himpl I]$};
\node (pa) at (20em,0em) {$B\himpl C$};
\node (pat) at (20.7em,0.44em) {};
\node[par] (pc) at (23em,1.732em) {};
\node (l) at (23em,1.732em) {\nodeindex{$\lambda$}};
\node (pb) at (26em,0.15em) {$B$};
\node (pd) at (23em,4.8em) {$C$};
\draw (pc) -- (pb);
\draw (pc) -- (pd);
\path[>=latex,->]  (pc) edge (pat);

\end{tikzpicture}
\end{center}
\caption{Links for HTLG proof structures}
\label{tab:lamlinks}
\end{table}

From Table~\ref{tab:lamlinks}, it is clear that par links have one premiss and two conclusions, whereas tensor links have two premisses and one conclusion (we will see a tensor link with one premiss and two conclusions later). Par links have an arrow pointing to the main formula of the link, the main formulas of tensor links are not distinguished visually (but can be determined from the formula labels). 

\begin{definition} A \emph{proof structure} is a tuple $\langle F, L\rangle$, where $F$ is a set of formula occurrences (vertices labeled with formulas) and $L$ is a set of links such that each link instantiates 
  one of the links of Table~\ref{tab:lamlinks}, and such that: 
\begin{itemize}
\item each formula is at most once the premiss of a link,
\item each formula is at most once the conclusion of a link.
\end{itemize}


The formulas which are not a conclusion of any link in a proof structure are its \emph{hypotheses}. We distinguish between \emph{lexical} hypotheses and \emph{logical} hypotheses; lexical hypotheses are formulas from the lexicon, all other hypotheses are logical. Conventionally all lexical hypotheses are written as strings rather than formulas. The formulas which are not a premiss of any link in a proof structure are its \emph{conclusions}. Formulas which are both a premiss and a conclusion of a link are \emph{internal nodes} of the proof structure.	

We say a proof structure with hypotheses $\Gamma$ and conclusions $\Delta$ is a proof structure of $\Gamma \vdash \Delta$, overloading the $\vdash$ symbol.
\end{definition}

\begin{definition} Given a proof structure $P$, a formula occurrence $A$ of $P$ is a \emph{cut formula} if it is the main formula of two links. $A$ is an \emph{axiomatic formula} in case it is not the main formula of any link. Formulas which are the main formula of exactly one link are \emph{flow formulas}.
\end{definition}

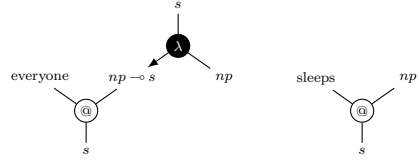
\begin{figure}
\begin{center}
\begin{tikzpicture}[scale=0.75]
\node (spar) at (3em,0em) {$s$};
\node (a) at (6em,4.8em) {$np\himpl s$};
\node (b) at (0em,4.8em) {everyone};
\node[tns] (c) at (3em,2.668em) {}; 
\node (cl) at (3em,2.668em) {$@$}; 
\draw (c) -- (spar);
\draw (c) -- (a);
\draw (c) -- (b);
\node (d) at (12em,4.8em) {$np$};
\node [par] (pc) at (9em,6.932em) {};
\node (l) at  (9em,6.932em) {\nodeindex{$\lambda$}};
\node (e) at (9em,9.6em) {$s$};
\path[>=latex,->]  (pc) edge (a);
\draw (pc) -- (d);
\draw (pc) -- (e);
\node (spar) at (21em,0em) {$s$};
\node (a) at (24em,4.8em) {$np$};
\node (b) at (18em,4.8em) {sleeps};
\node[tns] (c) at (21em,2.668em) {}; 
\node (cl) at (21em,2.668em) {$@$}; 
\draw (c) -- (spar);
\draw (c) -- (a);
\draw (c) -- (b);
\end{tikzpicture}
\end{center}
\caption{Proof structure of `everyone sleeps'.}
\label{fig:exps}
\end{figure}

\begin{example} As a very simple example, consider the lexicon containing only the words `everyone' of type $(np\himpl s)\himpl s$ with prosodic term $\lambda P. (P\,
\textit{everyone})$ and `sleeps' of type $np\himpl s$ with prosodic term $\lambda z. (z+\textit{sleeps})$. 
Unfolding the lexical entries produces the proof structure shown in Figure~\ref{fig:exps}. We use the convention of replacing lexical hypotheses with the corresponding word, so `everyone' represents the formula $(np\himpl s)\himpl s$ and `sleeps' the formula $np\himpl s$. The two other hypotheses of the proof structure (the $s$ premiss of the $[\himpl I]$ link and the $np$ premiss of the $[\himpl E]$ link) are logical hypotheses. There are no cut formulas in the figure, all complex formulas (the two lexical formulas and the $np\multimap s$ formula in the figure) are flow formulas,  all atomic formulas are axiomatic.

The figure shows a proof structure of $(np\himpl s)\himpl s, s, np\himpl s, np \vdash s, np, s$.
\end{example}

\editout{
\begin{definition}
Given a proof structure, $P = \langle v,l\rangle$, a \emph{substructure} $S= \langle v',l'\rangle$ is defined in one of three ways:
\begin{enumerate} 
\item \emph{delete links} we take a subset $d$ of $l$ 
 and set $l' = l \backslash d$ and $v' = v$, 
\item \emph{keep links} we take a subset $l'$ of $l$ and set $v'$ to be the union of all
neighboourhoods of $l'$,
\item \emph{keep vertices} we take a subset $v'$ of $v$ and add set $l'$ to all links of $l$ such that their neighbourhood $n$ is a subset of $v'$.
\end{enumerate}
\end{definition}
}

\begin{definition} Given a proof structure $P$ and two distinct formula occurrences $x$, $y$ of $P$, both labeled with the same formula $A$, with $x$ a logical hypothesis of $P$ and $y$ a conclusion of $P$. Then $P'$, the \emph{vertex contraction} (also called \emph{vertex identification}) of $x$ and $y$ in $P$, is the proof net obtained by deleting $x$ and $y$, adding a new node $z$ with label $A$ such that $z$ is the premiss of the link $x$ was a premiss of (if any) and the conclusion of the link that $y$ was the conclusion of (if any). 
\end{definition}

The vertex contraction operation is a standard graph theoretic operation \cite[p.\ 55]{bondymurty}. In the current context, it operates like the cut or axiom rule in the sense that if $P_1$ is a proof net of $\Gamma,A\vdash \Delta$ and $P_2$ a proof net of $\Gamma'\vdash A,\Delta'$ with $x$ and $y$ the two occurrences of $A$, then the vertex contraction of $x$ and $y$ is a proof net of $\Gamma,\Gamma' \vdash \Delta,\Delta'$. Given that, in an intuitionistic context like the current one, all proof nets have a single conclusion we even have that if $P_1$ is a proof net of $\Gamma,A\vdash C$ and $P_2$ a proof net of $\Gamma' \vdash A$, then the vertex contraction gives a proof net of $\Gamma,\Gamma' \vdash C$. Note that vertex contraction applies only to logical hypotheses and not to lexical ones.

Just like a logical link is a generalisation of a logical rule which is locally correct but need not be correct globally, a vertex contraction is a generalisation of the cut rule which is locally correct but need not be correct globally. In other words, the vertex contraction operation transforms proof structures to proof structures. Global correctness, as before, is verified by means of a correctness condition.
\moot{I've added some clarifying remarks. Is this better?} \symon{DO WE NEED TO CLARIFY LOCAL VERSION GLOBAL CORRECTNESS?}

\begin{example}
	Connecting the atomic formulas of the proof structure shown in Figure~\ref{fig:exps} produces the proof structure shown on the left of Figure~\ref{fig:expsb}. It has (the formulas corresponding to) `everyone' and `sleeps' as hypotheses (both lexical) and the formula $s$ as its conclusion, that is, it is a proof structure of $(np \himpl s) \himpl s, np\himpl s \vdash s$.
\end{example}

\begin{figure}
\begin{center}
\begin{tikzpicture}[scale=0.75]
\node (spar) at (3em,0em) {$s$};
\node (a) at (6em,4.8em) {$np\himpl s$};
\node (b) at (0em,4.8em) {everyone};
\node[tns] (c) at (3em,2.668em) {}; 
\node (cl) at (3em,2.668em) {$@$}; 
\draw (c) -- (spar);
\draw (c) -- (a);
\draw (c) -- (b);
%
\node [par] (pc) at (9em,6.932em) {};
\node (l) at  (9em,6.932em) {\textcolor{white}{$\lambda$}};
\node (e) at (9em,9.6em) {$s$};
\path[>=latex,->]  (pc) edge (a);
\draw (pc) -- (e);
\node (a) at (12em,14.4em) {$np$};
\node (b) at (6em,14.4em) {sleeps};
\node[tns] (c) at (9em,12.268em) {}; 
\node (cl) at (9em,12.268em) {$@$}; 
\draw (c) -- (e);
\draw (c) -- (a);
\draw (c) -- (b);
\draw (a) to [out=50,in=330] (pc);
\end{tikzpicture}
\begin{tikzpicture}[scale=0.75]
\node (ra) at (-6em,4.8em) {$\rightarrow$};
\node (raa) at (-6em,5.8em) {$\mathcal{A}$};
\node (spar) at (2em,0em) {$s$};
\node (a) at (6em,4.8em) {$\apsnodei$};
\node (ev) at (-2em,4.8em) {$\apsnodei$};
\node[tns] (c) at (2em,2.668em) {}; 
\node (cl) at (2em,2.668em) {$@$}; 
\draw (c) -- (spar);
\draw (c) -- (a);
\draw (c) -- (ev);
%
\node [par] (pc) at (9em,6.932em) {};
\node (l) at  (9em,6.932em) {\nodeindex{$\lambda$}};
\node (e) at (9em,9.6em) {$\apsnodei$};
\path[>=latex,->]  (pc) edge (a);
\draw (pc) -- (e);
\node (a) at (12em,14.4em) {$\apsnodei$};
\node (b) at (6em,15em) {$\apsnodei$};
\node[tns] (c) at (9em,12.268em) {}; 
\node (cl) at (9em,12.268em) {$@$}; 
\draw (c) -- (e);
\draw (c) -- (a);
\draw (c) -- (b);
\draw (a) to [out=50,in=330] (pc);
\node [tns] (pc) at (9em,17.668em) {};
\node (l) at  (9em,17.668em) {$\lambda$};
\draw (pc) -- (b);
\node (mid) at (9em,20.2em) {$\apsnodei$};
\draw (pc) -- (mid);
\node [tns] (tt) at (9em,22.468em) {};
\node (ttl) at (9em,22.468em) {$+$};
\draw (mid) -- (tt);
\node (left) at (6em,24.8em) {$\apsnodei$}; 
\node (leftt) at (6em,25.2em) {};
\node (right) at (12em,24.8em) {$\apsnodei$}; 
\draw (tt) -- (left);
\draw (tt) -- (right);
\draw plot [smooth, tension=1] coordinates { (leftt) (13em,27em) (13em,17em) (pc.south east)};
\node (sleeps) at (12em,25.5em) {\textit{sleeps}};
\node [tns] (pl) at (0em,6.932em) {};
\draw (pl) -- (ev);
\node (l) at  (0em,6.932em) {$\lambda$};
\node (e) at (0em,9.6em) {$\apsnodei$};
\draw (pl) -- (e);
\node (a) at (3em,14.4em) {$\apsnodei$};
\node (b) at (-3em,14.4em) {$\apsnodei$};
\node (p) at (-3em,14.8em) {};
\node[tns] (c) at (0em,12.268em) {}; 
\node (cl) at (0em,12.268em) {$@$}; 
\draw (c) -- (e);
\draw (c) -- (a);
\draw (c) -- (b);
%
%
\node (everyone) at (3em,15.1em) {\textit{everyone}};
\draw plot [smooth, tension=1] coordinates { (p) (4em,17.6em) (4em,7.6em) (pl.south east)};
\end{tikzpicture}
\end{center}
\caption{Proof structure of `everyone sleeps' after identification of the atomic formulas (left) and corresponding abstract proof structure (right).}
\label{fig:expsb}
\label{exaps}
\end{figure}

\begin{definition}
A \emph{tensor graph} is a connected proof structure with a unique conclusion (root) node containing only tensor links. The trivial tensor graph is a single node. 

Given a proof structure $P$, the \emph{components} of $P$ are the maximal substructures of $P$ which are tensor graphs. A \emph{tensor tree} is an acyclic tensor graph.
\end{definition}

\moot{TODO: the notion of tensor tree is probably superfluous in the current context and serves only to make the connection to multimodal proof nets; the notion of \emph{lambda graph} plays the same role in the current context as tensor trees do in the multimodal context.}
\moot{Correction to my previous remark: the notion of tensor tree remains useful for proof structures (e.g. the tensor case for the sequentialisation), whereas lambda graphs are useful for the abstract proof structure.}

For standard multimodal proof nets, we define correctness using tensor trees instead of the more general notion used here. Our results may be
(graph the\-o\-re\-ti\-cal representations of) lambda terms, and the $\lambda$ link represents the $\lambda$ binder for linear lambda terms. As is usual for lambda terms, we
need to be careful about `accidental capture' of variables. That is, we want to avoid incorrect reductions such as $(\lambda x \lambda y. (f\, x)) (g\, y)$ (not a linear
lambda term) to $\lambda y. (f\,(g\, y))$. \symon{WE DEFINE TENSOR TREES HERE BUT DO WE ACTUALLY USE THIS DEFINITION IN THIS SECTION?}\moot{See my TODO above, which used to be commented out.}

\begin{table}
\begin{center}
\begin{tikzpicture}[scale=0.75]
\node (labl) at (23em,-2.0em) {$[\lambda]$};
\node (pa) at (26em,0) {$\apsnodei$};
\node[tns] (pc) at (23em,1.732em) {};
\node (pcl) at (23em,1.732em) {$\lambda$};
\node (pb) at (20em,0.15em) {$\apsnodei$};
\node (pd) at (23em,4.8em) {$\apsnodei$};
\draw (pc) -- (pb);
\draw (pc) -- (pd);
\draw (pc) -- (pa);
\node (labl) at (-5em,-2.0em) {$[\epsilon]$};
\node (ab) at (-5em,0em) {$\apsnodei$};
\node (aba) at (-5em,0.3em) {};
\node[tns] (c) at (-5em,2.868em) {};
\node (plus) at (-5em,2.868em) {$\epsilon$};
\draw (c) -- (aba);
\node (labl) at (3em,-2.0em) {$[+]$};
\node (ab) at (3em,0em) {$\apsnodei$};
\node (aba) at (3em,0.3em) {};
\node (a) at (6em,4.8em) {$\apsnodei$};
\node (b) at (0em,4.8em) {$\apsnodei$};
\node[tns] (c) at (3em,2.868em) {};
\node (plus) at (3em,2.868em) {$+$};
\draw (c) -- (aba);
\draw (c) -- (a);
\draw (c) -- (b);
\node (labl) at (13em,-2.0em) {$[@]$};
\node (ab) at (13em,0em) {$\apsnodei$};
\node (aba) at (13em,0.3em) {};
\node (a) at (16em,4.8em) {$\apsnodei$};
\node (b) at (10em,4.8em) {$\apsnodei$};
\node[tns] (c) at (13em,2.868em) {};
\node (at) at (13em,2.868em) {$@$};
\draw (c) -- (aba);
\draw (c) -- (a);
\draw (c) -- (b);
\end{tikzpicture}

\begin{tikzpicture}[scale=0.75]
\node (aba) at (-5em,0.3em) {$\; \; \; \; $};
\node (labl) at (3em,-2.0em) {$[\ldr I]$};
\node (pa) at (0,0) {$\apsnodei$};
\node[par] (pc) at (3em,1.732em) {};
\node at (3em,1.732em) {\nodeindex{+}};
\node (pb) at (6em,0.15em) {$\apsnodei$};
\node (pd) at (3em,4.8em) {$\apsnodei$};
\draw (pc) -- (pb);
\draw (pc) -- (pd);
\path[>=latex,->]  (pc) edge (pa);
\node (labl) at (13em,-2.0em) {$[\ldl I]$};
\node (pa) at (16em,0) {$\apsnodei$};
\node[par] (pc) at (13em,1.732em) {}; 
\node at (13em,1.732em) {\nodeindex{+}};
\node (pb) at (10em,0em) {$\apsnodei$};
\node (pd) at (13em,4.8em) {$\apsnodei$};
\draw (pc) -- (pb);
\draw (pc) -- (pd);
\path[>=latex,->]  (pc) edge (pa);
\node (labl) at (23em,-2.0em) {$[\himpl I]$};
\node (pa) at (20em,0em) {$\apsnodei$};
\node (pat) at (20.7em,0.44em) {};
\node[par] (pc) at (23em,1.732em) {};
\node (l) at (23em,1.732em) {\nodeindex{$\lambda$}};
\node (pb) at (26em,0.15em) {$\apsnodei$};
\node (pd) at (23em,4.8em) {$\apsnodei$};
\draw (pc) -- (pb);
\draw (pc) -- (pd);
\path[>=latex,->]  (pc) edge (pa);
\end{tikzpicture}

\end{center}
\caption{Links for HTLG abstract proof structures}
\label{tab:apslinks}
\end{table}

\subsection{Abstract proof structures}

For proof nets, correctness is defined on graph theoretic representations obtained from proof structures by forgetting some of the formula labels. We call these representations \emph{abstract proof structures}. A more procedural way of seeing abstract proof structures is as a way of computing the structure of the antecedent. For hybrid type-logical grammars, this means abstract proof structures must contain some way of representing lambda terms in addition to the Lambek calculus structures.

\begin{definition}
An abstract proof structure $\mathcal{A}$ is a tuple $\langle V, L, l, h, c\rangle$ where $V$ is a set of vertices, $L$ is a set of the links shown in Table~\ref{tab:apslinks} connecting the vertices of $V$, $l$ is a function from the lexical hypothesis vertices of $\mathcal{A}$ to the corresponding variables, $h$ is a function from logical hypothesis vertices to formulas, and $c$ is a function from the conclusion vertices of $\mathcal{A}$ to formulas (a hypothesis vertex is a vertex which is not the conclusion of any link in $L$,  and a conclusion vertex is a vertex which is not the premiss of any link in $L$). 	
\end{definition}


The links for abstract proof structures are shown in Table~\ref{tab:apslinks}. The tensor links are shown in the topmost row, the par links in the bottom row, with the par links for the Lambek connectives on the left and in the middle, and the par link for the linear implication on the bottom right.

The $\lambda$ tensor link is the only non-standard link. Even though it has the same shape as the link for the Grishin  connectives of \citet{mm12pnlg}, it is used in a rather different way. The $\lambda$ tensor link does not correspond to a logical connective but rather to lambda abstraction over variables (or rather their graph theoretical representations). To keep the logic simple and the number of connectives as small as possible, we have chosen to make the $\epsilon$ link, corresponding to the empty string, a non-logical link as well. As a consequence, $\epsilon$ can appear only in lexical terms. However, if needed, it would be easy to adapt the logic by adding a logical connective $1$ corresponding to $\epsilon$.



\begin{definition}
Given a proof structure $P$, we obtain the corresponding abstract proof structure $\textit{aps}(P) = \mathcal{A}$ as follows.
\begin{enumerate}
\item we keep the set of vertices $V$ and the set of links $L$ of $P$ (but we forget the formula labels of the internal nodes), 
\item logical hypotheses are kept as simple vertices,
but we replace each lexical hypothesis $M:A$ of the proof structure by a graph $g$ corresponding to its lambda term $M$, the conclusion of $g$ is the vertex which was the lexical hypothesis of $P$, making the word subterm $w$ of $M$ a lexical hypothesis of the new structure, 
\item we define $l$ to assign the corresponding word for each lexical hypothesis of the resulting graph, $h$ to assign a formula for all logical hypotheses, and $c$ to assign a formula to all conclusions.
\end{enumerate}
\end{definition}

\begin{example}

Converting the proof structure on the left of Figure~\ref{fig:expsb} to an abstract proof structure produces the abstract proof structure shown on the right.
The three links from the proof structure have been preserved: only their internal formulas have been removed. We have replaced `everyone', corresponding to the formula $(np\himpl s) \himpl s$, by the graph structure corresponding to its lexical lambda term $\lambda P. (P\,\textit{everyone})$ and similarly for `sleeps', corresponding to formula $np\himpl s$, which we have replaced by the graph corresponding to $\lambda x. x+\textit{sleeps}$.

The shift from proof structure to abstract proof structure corresponds to a shift from the formula level to the term level. The $\lambda$ par link in the proof structure corresponds to the introduction rule (once we have verified its correct application by means of the correctness condition) whereas its occurrence in the abstract proof structure corresponds to abstraction (again modulo the correctness condition). Similarly, the two tensor links in the proof structure correspond to the elimination rule for linear application whereas the corresponding links in the abstract proof structure correspond to application.
The goal formula $s$ represents the root node of the term to be computed.
\symon{ Should there be a brief but more pedagogical explanation of why the abstract proof structure contains several more links than the proof structure in Figure 8? This is mentioned in step two of the translation function, but here's one explanation of the size of abstract proof structures: The abstract proof structure may be larger than the input proof structure--as it is here--since it represents the prosodic lambda terms of the lexical formulae, as noted in the second point of Definition 8. This information is necessary to ensure ensure correctness. Thus, for instance, the variable P in $\lambda P.P(everyone)$ composes with the string \textit{everyone} via an `@' link and is subsequently withdrawn via the `$\lambda$' link. This graph, conditions on which are defined below, is then the abstract proof structure corresponding to the lexical formula \textit{everyone} in the input proof structure.}

\end{example}

\begin{definition} A \emph{lambda graph} is an abstract proof structure such that:
\begin{enumerate}
    \item it has a single conclusion,
	\item it contains only tensor links,
	\item\label{item:linearlambda} each right conclusion of a lambda link is an ancestor of its premiss,
	\item\label{item:acc} removing the connection between all lambda links and their rightmost conclusion produces an acyclic and connected structure. 
\end{enumerate}	
\end{definition}

Condition~\ref{item:linearlambda} avoids vacuous abstraction and accidental variable capture in the corresponding lambda term. Condition~\ref{item:acc} is the standard acyclicity and connectedness condition for abstract proof structures, but allowing for the fact that lambda abstraction (but no other tensor links) can produce cycles.





Lambda graphs correspond to linear lambda terms in the obvious way, with the rightmost conclusion of the lambda link representing the variable abstracted over. This is a standard way of representing lambda terms in a way which avoids the necessity of variable renaming (alpha conversion). 


\begin{proposition}
a lambda term with free variables $x_1,\ldots,x_n$ corresponds to a lambda graph with hypotheses $x_1,\ldots,x_n$, with the $@$ tensor link corresponding to application, the $\lambda$ tensor link to abstraction, and the $+$ link and the $\epsilon$ link to the term constants of type $s\rightarrow s\rightarrow s$ and $s$ respectively. To keep the terms simple, we will write $(X + Y)$ instead of $((+ X) Y)$. 
\end{proposition}

\subsection{Structural rules and contractions}

To decide whether or not a given proof structure is a \emph{proof net} (that is, corresponds to a natural deduction proof), we will introduce a system of graph rewriting. The structural rules for the non-associative version of hybrid type-logical grammars are shown on the left-hand column of Table~\ref{tab:truesrlam}. We can obtain the standard associative version simply by adding the associativity rules for `$+$'; more generally, we can add any structural conversion for multimodal grammars, rewriting a tensor tree into another tensor tree with the same leaves (though not necessarily in the same order) provided they do not overlap with the beta redex. The $\epsilon$ structural rules simply stipulate that `$\epsilon$' functions as the identity element for `$+$' (both as a left identity and as a right identity). 

The key rewrite is the beta conversion rule. It is the graph theoretical equivalent of performing a beta reduction on the corresponding term. For the beta rewrite, we
replace the two links (and the internal node) and perform two vertex contractions: $h_1$ with $c_1$ and $h_2$ with $c_2$. We update the functions $h$, $l$ and $c$
accordingly (if one of the $h_i$ was in the domain of $h$ and $l$ then so is the resulting vertex, and similarly for the $c_i$ and the $c$ function of the abstract proof structure).


\begin{table}
\begin{center}
\begin{tabular}{ccc}
\begin{tikzpicture}[scale=0.75]
\node at (28em,-8em) {$\apsnode{h_1}{c_1}$};
\node at (28em,-14em) {$\apsnode{h_2}{c_2}$}; 
\node at (24.0em,-11em) {$\rightarrow$}; 
\node at (24.0em,-10em) {$[\beta]$};
\node at (24.0em,-2em) {\emph{\large{Structural rules}}}; 
%
\node (lt) at (19em,-13.5em) {$\apsnode{h_1}{}$}; 
\node[tns] (tc) at (16em,-15.232em) {}; 
\node (appl) at (16em,-15.232em) {$@$};
\node (rt) at (13em,-12.5em) {$\apsnode{}{}$}; 
\node (top) at (16em,-18.3em) {$\apsnode{}{c_2}$}; 
\draw (lt) -- (tc);
\draw (rt) -- (tc);
\draw (tc) -- (top);
\node[tns] (lam) at (16em,-9.232em) {}; 
\node (lamlab) at (16em,-9.232em) {$\lambda$}; 
\node (bl) at (16em,-6.7em) {$\apsnode{h_2}{}$};
\node (bll) at (16em,-6.7em) {\phantom{M}}; 
\draw (bll) -- (lam);
\draw (lam) -- (rt);
\node (bot) at (19em,-10.5em) {$\apsnode{}{c_1}$};
\draw (lam) -- (bot); 
\end{tikzpicture}
& \qquad\qquad &
\begin{tikzpicture}[scale=0.75]
\node (ab) at (3em,4.8em) {$\apsnodei$};
\node (a) at (0,9.6em) {$\apsnode{h}{}$};
\node (b) at (6em,9.6em) {$\apsnodei$};
\node[tns] (c) at (3em,7.668em) {};
\node (cl) at (3em,7.668em) {$+$};
\draw (c) -- (ab);
\draw (c) -- (a);
\draw (c) -- (b);
\node (pa) at (0,0) {$\apsnode{}{c}$};
\node[par] (pc) at (3em,1.732em) {};
\node at (3em,1.732em) {\nodeindex{+}};
\draw (pc) -- (ab);
\path[>=latex,->]  (pc) edge (pa);
\draw (b) to [out=50,in=330] (pc);
%
%
\node at (14em,5.0em) {$\apsnode{h}{c}$}; 
\node (labl) at (10em,6.0em) {$[\ldr I]$}; 
\node  at (10em,5.0em) {$\rightarrow$}; 
\node at (10em,16em) {\emph{\large{Logical contractions}}};
\node (align) at (-2em,12em) {\phantom{M}};
\end{tikzpicture}
\\
\begin{tikzpicture}[scale=0.75]
\node (ab) at (3em,0em) {$\apsnode{}{c}$};
\node (aba) at (3em,0.3em) {};
\node (aa) at (6em,4.8em) {\phantom{M}};
\node (a) at (6em,4.8em) {$\apsnode{h}{}$};
\node (b) at (0em,4.8em) {$\apsnodei$};
\node[tns] (c) at (3em,2.868em) {};
\node (plus) at (3em,2.868em) {$+$};
\draw (c) -- (aba);
\draw (c) -- (a);
\draw (c) -- (b);
\node[tns] (eps) at (0em,7.2em) {$\epsilon$};
\draw (b) -- (eps);
\node (rhs) at (14em,3em) {$\apsnode{h}{c}$};
\node at (10em,3em) {$\rightarrow$}; 
\node at (10em,4em) {$[\epsilon_L]$}; 
\end{tikzpicture}
& \qquad & 
\begin{tikzpicture}[scale=0.75]
\node (ab) at (3em,4.8em) {$\apsnodei$};
\node (a) at (6em,9.6em) {$\apsnode{h}{}$};
\node (b) at (0em,9.6em) {$\apsnodei$};
\node[tns] (c) at (3em,7.668em) {};
\node (cl) at (3em,7.668em) {$+$};
\draw (c) -- (ab);
\draw (c) -- (a);
\draw (c) -- (b);
\node (pa) at (6em,0) {$\apsnode{}{c}$};
\node[par] (pc) at (3em,1.732em) {};
\node at (3em,1.732em) {\nodeindex{+}};
\draw (pc) -- (ab);
\path[>=latex,->]  (pc) edge (pa);
\draw (b) to [out=130,in=210] (pc);
\node at (14em,5.0em) {$\apsnode{h}{c}$}; 
\node (labl) at (10em,6.0em) {$[\ldl I]$}; 
\node  at (10em,5.0em) {$\rightarrow$}; 
\end{tikzpicture}
\\
\begin{tikzpicture}[scale=0.75]
\node (ab) at (3em,0em) {$\apsnode{}{c}$};
\node (aba) at (3em,0.3em) {};
\node (aa) at (0em,4.8em) {\phantom{M}};
\node (a) at (0em,4.8em) {$\apsnode{h}{}$};
\node (b) at (6em,4.8em) {$\apsnodei$};
\node[tns] (c) at (3em,2.868em) {};
\node (plus) at (3em,2.868em) {$+$};
\draw (c) -- (aba);
\draw (c) -- (a);
\draw (c) -- (b);
\node[tns] (eps) at (6em,7.2em) {$\epsilon$};
\draw (b) -- (eps);
\node (rhs) at (14em,3em) {$\apsnode{h}{c}$};
\node at (10em,3em) {$\rightarrow$}; 
\node at (10em,4em) {$[\epsilon_R]$}; 
\end{tikzpicture}
& \qquad &
\begin{tikzpicture}[scale=0.75]
\node at (20em,2em) {$\rightarrow$};
\node (labl) at (20em,3.0em) {$[\himpl I]$};
\node (pa) at (10em,0em) {$\apsnode{}{c_1}$};
\node (pat) at (10.7em,0.44em) {};
\node[par] (pc) at (13em,1.732em) {};
\node (l) at (13em,1.732em) {\nodeindex{$\lambda$}};
\node (pb) at (16em,0.0em) {$\apsnode{}{c_2}$};
\node (pd) at (13em,4.8em) {$\apsnode{h}{}$};
\node (pdd) at (13em,4.8em) {\phantom{M}};
\draw (pc) -- (pb);
\draw (pc) -- (pdd);
\path[>=latex,->]  (pc) edge (pa);
\node (pa) at (24em,0em) {$\apsnode{}{c_1}$};
\node (pat) at (24.7em,0.44em) {};
\node[tns] (pc) at (27em,1.732em) {};
\node (l) at (27em,1.732em) {$\lambda$};
\node (pb) at (30em,0.0em) {$\apsnode{}{c_2}$};
\node (pd) at (27em,4.8em) {$\apsnode{h}{}$};
\node (pdd) at (27em,4.8em) {\phantom{M}};
\draw (pc) -- (pb);
\draw (pc) -- (pdd);
\draw (pc) -- (pa);
\node (align) at (2em,8em) {\phantom{M}};
\end{tikzpicture}
\end{tabular}
\end{center}
\caption{Structural rules (left) and logical contractions (right) for HTLG proof nets.}
\label{tab:truesrlam}
\label{tab:contr}
\end{table}

To make the operation of the beta reduction clearer, a `sugared' version of the contraction is shown in Table~\ref{tab:srlamterm}. Term labels have been added to the vertices of the graph to make the correspondence with beta-reduction explicit.
This second picture is slightly misleading in that it suggests that $A_1$, $A_2$ and $A_3$ are disjoint substructures. This need not be the case: for example, $A_3$ can contain a lambda link whose right conclusion is a premiss of either $A_1$ or $A_2$. Similarly, in a logic with the Lambek calculus product, the link for $[\bullet E]$ may connect premisses of both $A_1$ and $A_2$. A side condition on the $\himpl I$ conversion combined with the restriction of lexical entries to linear lambda terms will guarantee that $x$ ($c_1$) in the beta reduction is always a descendant of $N$($h_2$). 

\begin{table}
\begin{center}
\begin{tikzpicture}[scale=0.75]
\node[pn] at (33em,-11em) {$\mathcal{A}_2$}; 
\node[pn] at (33em,-5.0em) {$\mathcal{A}_1$}; 
\node[pn] at (33em,-17.0em) {$\mathcal{A}_3$}; 
\node at (33em,-8em) {$M$};
\node at (33em,-14em) {$N[x:=M]$}; 
\node at (26.5em,-11em) {$\rightarrow$}; 
\node at (26.5em,-10em) {$[\beta]$}; 
%
\node (lt) at (19em,-13.5em) {$M$}; 
\node[tns] (tc) at (16em,-15.232em) {}; 
\node (appl) at (16em,-15.232em) {$@$};
\node (rt) at (13em,-13.5em) {$\lambda x. N$}; 
\node (top) at (16em,-18.3em) {$(\lambda x. N)\, M$}; 
\draw (lt) -- (tc);
\draw (rt) -- (tc);
\draw (tc) -- (top);
\node[tns] (lam) at (16em,-10.232em) {}; 
\node (lamlab) at (16em,-10.232em) {$\lambda$}; 
\node (bl) at (16em,-7.7em) {$N$}; 
\draw (bl) -- (lam);
\draw (lam) -- (rt);
\node[pn] at (15.5em,-4.8em) {$\mathcal{A}_2$}; 
\node[pn] at (21.0em,-10.7em) {$\mathcal{A}_1$}; 
\node[pn] at (16.0em,-21.3em) {$\mathcal{A}_3$}; 
\node (bot) at (16em,-2.0em) {$x$}; 
\draw (lam) to [out=-50,in=-330] (bot); 
\end{tikzpicture}
\end{center}
\caption{Beta conversion as a structural rule with term labels added.}
\label{tab:srlamterm}
\end{table}

\begin{definition} We say a lambda graph is \emph{normal} or $beta$-\emph{normal} when it doesn't contain any redexes for the beta conversion.
\end{definition}

In addition to the structural rules, there are contractions for each of the logical connectives. Table~\ref{tab:contr} shows, on the right-hand column, the contractions for HTLG.
For the Lambek implications, these are just the standard
contractions. They combine a concatenation mode `+' with one of its residuals\footnote{To ensure confluence of `$/$' and `$\backslash$' in the presence of $\epsilon$ we can add the side condition to the $[/I]$ and $[\backslash I]$ contractions that the component to which the par link is attached has at least one hypothesis other than the	 auxiliary conclusion of the par link. This forbids empty antecedent derivations and restores confluence.}. The contractions for the Lambek calculus implications are the standard contractions from \citet{mp}.

The contraction for $\himpl I$ has the side condition that the rightmost conclusion of the $\lambda$ par link is a descendant of its premiss, passing
only through tensor links. This is essentially the same condition as the one used by \citet{reductions}, only without performing the actual contractions. This is because we want our
abstract proof structures to represent the prosodic structure of a proof, which may contain lambda terms, just like the standard goal of abstract proof structures is
always to compute the structure which would make the derivation valid. \symon{COULD WE ALSO SAY THAT WE DON'T WANT TO CONTRACT THE $\multimap$ I VERTICES BECAUSE DOING SO WOULD}\moot{There seems to be a part missing here. Let me know what you would like to add here for clarification.}

Our rewrite calculus can be situated in the larger context of adding rewrite rules to the lambda calculus \cite{klop93crs,ck96alglambda}. Even though the contractions for $[/I]$ and $[\backslash I]$  are not left-linear, since they correspond to terms $(M+x)/x$  and $x\backslash (x+M)$ respectively, this is not a problem because the occurrences of $x$ are bound occurrences \cite{klop93crs}. In general, confluence can not be maintained in the presence of structural rules (or of the unary connectives) since the structural rules themselves need not be confluent. Confluence of beta reduction is guaranteed by not allowing any structural rewrite to overlap with the beta redex \cite{klop93crs}.  

\subsection{The eta rule and associativity}
\label{sec:eta}
\label{sec:ass}

We can add an addition structural rule corresponding to eta conversion in the lambda calculus as shown in Table~\ref{tab:eta}.

\begin{table}
\begin{center}
\begin{tikzpicture}[scale=0.75]
\node (ab) at (3em,4.8em) {$\apsnodei$};
\node (a) at (0,9.6em) {$\apsnode{h}{}$};
\node (b) at (6em,9.6em) {$\apsnodei$};
\node[tns] (c) at (3em,7.668em) {};
\node (cl) at (3em,7.668em) {$@$};
\draw (c) -- (ab);
\draw (c) -- (a);
\draw (c) -- (b);
\node (pa) at (0,0) {$\apsnode{}{c}$};
\node[tns] (pc) at (3em,1.732em) {};
\node at (3em,1.732em) {$\lambda$};
\draw (pc) -- (ab);
\path[>=latex,->]  (pc) edge (pa);
\draw (b) to [out=50,in=330] (pc);
\node (labl) at (10em,6.0em) {$[\eta]$}; 
\node  at (10em,5.0em) {$\rightarrow$}; 
\node (align) at (-2em,12em) {\phantom{M}};
\node at (14em,5.0em) {$\apsnode{h}{c}$}; 
\end{tikzpicture}
\end{center}
\caption{The eta conversion as a structural rule}
\label{tab:eta}
\end{table}

The eta reduction in the lambda calculus converts $\lambda x. (P\,x)$ to $P$, with the condition that $P$ does not contain occurrences of $x$. In linear lambda terms, $(P x)$ must contain exactly one occurence of $x$ by the linearity condition on lambda terms (since the full term is $\lambda x. (P\,x)$), therefore $P$ cannot contain any occurrences of $x$, and this condition is automatically satisfied.

\begin{figure}
  \begin{center}
\begin{tikzpicture}[scale=0.75]
\node (ab) at (3em,4.8em) {$A$};
\node (a) at (0,9.6em) {$A\himpl B$};
\node (b) at (6em,9.6em) {$B$};
\node[tns] (c) at (3em,7.668em) {};
\node (cl) at (3em,7.668em) {$@$};
\draw (c) -- (ab);
\draw (c) -- (a);
\draw (c) -- (b);
\node (pa) at (0,0) {$A\himpl B$};
\node[par] (pc) at (3em,1.732em) {};
\node at (3em,1.732em) {\nodeindex{$\lambda$}};
\draw (pc) -- (ab);
\path[>=latex,->]  (pc) edge (pa);
\draw (b) to [out=50,in=330] (pc);
\node (labl) at (10em,6.0em) {$\mathcal{A}$}; 
\node  at (10em,5.0em) {$\rightarrow$}; 
\end{tikzpicture}
\begin{tikzpicture}[scale=0.75]
\node (ab) at (3em,4.8em) {$\apsnodei$};
\node (a) at (0,9.6em) {$\smash{\apsnode{A\himpl B}{}}$};
\node (b) at (6em,9.6em) {$\apsnodei$};
\node[tns] (c) at (3em,7.668em) {};
\node (cl) at (3em,7.668em) {$@$};
\draw (c) -- (ab);
\draw (c) -- (a);
\draw (c) -- (b);
\node (pa) at (0,0) {$\smash{\apsnode{}{A\himpl B}}$};
\node[par] (pc) at (3em,1.732em) {};
\node at (3em,1.732em) {\nodeindex{$\lambda$}};
\draw (pc) -- (ab);
\path[>=latex,->]  (pc) edge (pa);
\draw (b) to [out=50,in=330] (pc);
\node (labl) at (10em,6.0em) {$[\himpl I]$}; 
\node  at (10em,5.0em) {$\rightarrow$}; 
\end{tikzpicture}
\!\!\!\!\!\!\!\!\!\!\!
\begin{tikzpicture}[scale=0.75]
\node (ab) at (3em,4.8em) {$\apsnodei$};
\node (a) at (0,9.6em) {$\smash{\apsnode{A\himpl B}{}}$};
\node (b) at (6em,9.6em) {$\apsnodei$};
\node[tns] (c) at (3em,7.668em) {};
\node (cl) at (3em,7.668em) {$@$};
\draw (c) -- (ab);
\draw (c) -- (a);
\draw (c) -- (b);
\node (pa) at (0,0) {$\smash{\apsnode{}{A\himpl B}}$};
\node[tns] (pc) at (3em,1.732em) {};
\node at (3em,1.732em) {$\lambda$};
\draw (pc) -- (ab);
\path[>=latex,->]  (pc) edge (pa);
\draw (b) to [out=50,in=330] (pc);
\node (labl) at (10em,6.0em) {$[\eta]$}; 
\node  at (10em,5.0em) {$\rightarrow$}; 
\node (align) at (-2em,12em) {\phantom{M}};
\node at (14em,5.0em) {$\smash{\apsnode{A\himpl B}{A\himpl B}}$}; 
\end{tikzpicture}
\end{center}
\caption{An example eta conversion}
\label{fig:eta}
\end{figure}
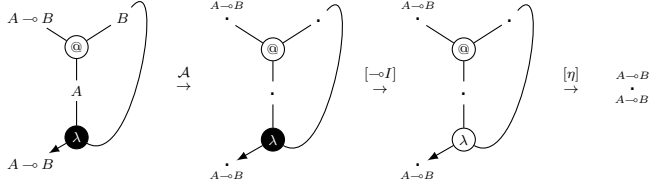

Figure~\ref{fig:eta} shows an example of eta reduction, with the proof structure on the left, then the corresponding abstract proof structure, the $\multimap I$ conversion, and finally the $\eta$ conversion. The $\multimap I$ and $\eta$ conversion together function for application and abstraction as the $\ldr I$ contraction does for concatenation.

Hybrid type-logical grammars are generally presented as an associative calculus, whereas so far we have studied a non-associative version of the logic. We can make the logic associative simply by adding the structural rules for associativity of `$+$' to the natural deduction calculus, sequent calculus and the proof net calculus. However, explicitly rebracketing structures is a source of inefficiency, and we generally prefer a notation which leaves the brackets implicit. For the proof net calculus, we can follow \citet{mp} and use an $n$-premiss version of the tensor link for $+$. This gives the structural rule and contractions shown in Table~\ref{tab:ass}.

\begin{table}
\begin{center}
\begin{tikzpicture}[scale=0.75]
\node (mid) at (3em,4.8em) {$\apsnode{}{c}$};
\node (a) at (0em,9.6em) {$\apsnode{h_1}{}$};
\node (ab) at (1.5em,9.6em) {$\ldots$};
\node (ad) at (3em,9.6em) {$\apsnodei$};
\node (ab) at (4.5em,9.6em) {$\ldots$};
\node (ax) at (0em,7.668em) {};
\node (bx) at (6em,7.668em) {};
\node (b) at (6em,9.6em) {$\apsnode{h_{n+m}}{}$};
\node[tns] (c) at (3em,7.668em) {};
\node (cl) at (3em,7.668em) {$+$};
\draw (c) -- (mid);
\draw (c) -- (0em,7.668em);
\draw (0em,7.668em) -- (0em,9em);
\draw (c) -- (6em,7.668em) -- (6em,9em);
\draw (1.5em,7.668em) -- (1.5em,9em);
\draw (4.5em,7.668em) -- (4.5em,9em);
\draw (c) -- (3em,9em);
\node[tns] (cc) at (3em,12.468em) {};
\node (ccl) at (3em,12.468em) {$+$};
\draw (cc) -- (ad);
\node (hk) at (1.5em,14.4em) {$\apsnode{h_k}{}$};
\node (hknm) at (4.5em,14.4em) {$\apsnode{h_{k+m-1}}{}$};
\node (td) at (3em,14.4em) {$\ldots$};
\draw (cc) -- (1.5em,12.468em) -- (1.5em,13.8em);
\draw (cc) -- (4.5em,12.468em) -- (4.5em,13.8em);
\draw (cc) -- (3em,13.8em);
%
%
\node (spc) at (10em,2.4em) {};
\node (labl) at (10em,6.0em) {$[\textit{Ass}]$}; 
\node  at (10em,5.0em) {$\rightarrow$}; 
\node (mid) at (18em,4.8em) {$\apsnode{}{c}$};
\node (a) at (13.5em,9.6em) {$\apsnode{h_1}{}$};
\node (ad) at (15em,9.6em) {$\ldots$};
\node (ab) at (16.5em,9.6em) {$\apsnode{h_k}{}$};\%
\node (ad) at (18em,9.6em) {$\ldots$};
\node (ab) at (19.5em,9.6em) {$\apsnode{h_{k+m-1}}{}$};
\node (ad) at (21em,9.6em) {$\ldots$};
\node (ab) at (22.5em,9.6em) {$\apsnode{h_{n+m}}{}$};
\node[tns] (c) at (18em,7.668em) {};
\node (cl) at (18em,7.668em) {$+$};
\draw (c) -- (mid);
\draw (c) -- (13.5em,7.668em);
\draw (c) -- (22.5em,7.668em) -- (22.5em,9em);
\draw (13.5em,7.668em) -- (13.5em,9em);
\draw (c) -- (18em,9em);
\draw (15em,7.668em) -- (15em,9em);
\draw (16.5em,7.668em) -- (16.5em,9em);
\draw (21em,7.668em) -- (21em,9em);
\draw (19.5em,7.668em) -- (19.5em,9em);
\end{tikzpicture}

\begin{tabular}{ccc}
\begin{tikzpicture}[scale=0.75]
\node (mid) at (3em,4.8em) {$\apsnodei$};
\node (a) at (0em,9.6em) {$\apsnode{h_1}{}$};
\node (ab) at (1.5em,9.6em) {$\apsnode{h_2}{}$};
\node (ad) at (3em,9.6em) {$\ldots$};
\node (ab) at (4.5em,9.6em) {$\apsnode{h_n}{}$};
\node (ax) at (0em,7.668em) {};
\node (bx) at (6em,7.668em) {};
\node (b) at (6em,9.6em) {$\apsnodei$};
\node[tns] (c) at (3em,7.668em) {};
\node (cl) at (3em,7.668em) {$+$};
\draw (c) -- (mid);
\draw (c) -- (0em,7.668em);
\draw (0em,7.668em) -- (0em,9em);
\draw (c) -- (6em,7.668em) -- (6em,9em);
\draw (1.5em,7.668em) -- (1.5em,9em);
\draw (4.5em,7.668em) -- (4.5em,9em);
\draw (c) -- (3em,9em);
\node (pa) at (0,0) {$\apsnode{}{c}$};
\node[par] (pc) at (3em,1.732em) {};
\node at (3em,1.732em) {\nodeindex{+}};
\draw (pc) -- (mid);
\path[>=latex,->]  (pc) edge (pa);
\draw (b) to [out=50,in=330] (pc);
%
%
%
\node (labl) at (10em,6.0em) {$[\ldr I]$}; 
\node  at (10em,5.0em) {$\rightarrow$}; 
\node (mid) at (15em,4.8em) {$\apsnode{}{c}$};
\node (a) at (12em,9.6em) {$\apsnode{h_1}{}$};
\node (ab) at (14em,9.6em) {$\apsnode{h_2}{}$};
\node (ad) at (16em,9.6em) {$\ldots$};
\node (ab) at (18em,9.6em) {$\apsnode{h_n}{}$};
\node[tns] (c) at (15em,7.668em) {};
\node (cl) at (15em,7.668em) {$+$};
\draw (c) -- (mid);
\draw (c) -- (12em,7.668em);
\draw (12em,7.668em) -- (12em,9em);
\draw (c) -- (18em,7.668em) -- (18em,9em);
\draw (14em,7.668em) -- (14em,9em);
\draw (16em,7.668em) -- (16em,9em);
\end{tikzpicture}
& \qquad &
\begin{tikzpicture}[scale=0.75]
\node (ab) at (3em,4.8em) {$\apsnodei$};
\node (b) at (0em,9.6em) {$\apsnodei$};
\node[tns] (c) at (3em,7.668em) {};
\node (cl) at (3em,7.668em) {$+$};
%
\node (pa) at (6em,0) {$\apsnode{}{c}$};
\node[par] (pc) at (3em,1.732em) {};
\node at (3em,1.732em) {\nodeindex{+}};
\draw (pc) -- (ab);
\path[>=latex,->]  (pc) edge (pa);
\draw (b) to [out=130,in=210] (pc);
\node (mid) at (3em,4.8em) {$\apsnodei$};
\node (aa) at (0em,9.6em) {$\apsnodei$};
\node (a) at (1.5em,9.6em) {$\apsnode{h_1}{}$};
\node (ab) at (3.0em,9.6em) {$\apsnode{h_2}{}$};
\node (ad) at (4.5em,9.6em) {$\ldots$};
\node (ab) at (6.0em,9.6em) {$\apsnode{h_n}{}$};
\node (ax) at (0em,7.668em) {};
\node (bx) at (6em,7.668em) {};
\node[tns] (c) at (3em,7.668em) {};
\node (cl) at (3em,7.668em) {$+$};
\draw (c) -- (mid);
\draw (c) -- (0em,7.668em);
\draw (0em,7.668em) -- (0em,9em);
\draw (c) -- (6em,7.668em) -- (6em,9em);
\draw (1.5em,7.668em) -- (1.5em,9em);
\draw (4.5em,7.668em) -- (4.5em,9em);
\draw (c) -- (3em,9em);
%
\node[par] (pc) at (3em,1.732em) {};
\node at (3em,1.732em) {\nodeindex{+}};
\draw (pc) -- (mid);
%
%
%
\node (labl) at (10em,6.0em) {$[\ldl I]$}; 
\node  at (10em,5.0em) {$\rightarrow$}; 
\node (mid) at (15em,4.8em) {$\apsnode{}{c}$};
\node (a) at (12em,9.6em) {$\apsnode{h_1}{}$};
\node (ab) at (14em,9.6em) {$\apsnode{h_2}{}$};
\node (ad) at (16em,9.6em) {$\ldots$};
\node (ab) at (18em,9.6em) {$\apsnode{h_n}{}$};
\node[tns] (c) at (15em,7.668em) {};
\node (cl) at (15em,7.668em) {$+$};
\draw (c) -- (mid);
\draw (c) -- (12em,7.668em);
\draw (12em,7.668em) -- (12em,9em);
\draw (c) -- (18em,7.668em) -- (18em,9em);
\draw (14em,7.668em) -- (14em,9em);
\draw (16em,7.668em) -- (16em,9em);
\end{tikzpicture}
\end{tabular}
\end{center}
\caption{Structural rule and contractions for associativity}
\label{tab:ass}
\end{table}

The associativity rule simply combines an $m$-premiss link with an $n$-premiss link to produce an $n+m$-premiss link, preserving the linear order of the premisses. Repeated application of the associativity rule replaces a binary tree of concatenation links by a single link with the yield of the initial tree as its hypotheses.  To be fully precise, the $\epsilon$ structural rules should be updated as well in the new format (alternatively, we can treat it as the special case of the associativity rule with $m=0$, that is, with the topmost link being a zero-premiss link).

The $\ldl I$ and $\ldr I$ links remove the leftmost and rightmost premiss of an $n+1$ premiss link (given the empty antecedent restriction, $n$ should be greater than 0).

Compared to explicit associative tree rebracketing rules, the rules of Table~\ref{tab:ass} have the advantage that they reduce the size of the structure, and are therefore easier to use for complexity analysis and computational implementation.

\subsection{Correctness of the proof net calculus}
\label{sec:correct}

\begin{definition}
A proof structure is a \emph{proof net} whenever its abstract proof structure converts to a lambda graph. 
\end{definition}

\begin{figure}
\begin{center}
\begin{tikzpicture}[scale=0.75]
\node (spar) at (2em,0em) {$s$};
\node (a) at (6em,4.8em) {$\apsnodei$};
\node (ev) at (-2em,4.8em) {$\apsnodei$};
\node[tns] (c) at (2em,2.668em) {}; 
\node (cl) at (2em,2.668em) {$@$}; 
\draw (c) -- (spar);
\draw (c) -- (a);
\draw (c) -- (ev);
%
\node [par] (pc) at (9em,6.932em) {};
\node (l) at  (9em,6.932em) {\nodeindex{$\lambda$}};
\node (e) at (9em,9.6em) {$\apsnodei$};
\path[>=latex,->]  (pc) edge (a);
\draw (pc) -- (e);
\node (a) at (12em,14.4em) {$\apsnodei$};
\node (b) at (6em,15em) {$\apsnodei$};
\node[tns] (c) at (9em,12.268em) {}; 
\node (cl) at (9em,12.268em) {$@$}; 
\draw (c) -- (e);
\draw (c) -- (a);
\draw (c) -- (b);
\draw (a) to [out=50,in=330] (pc);
\node [tns] (pc) at (9em,17.668em) {};
\node (l) at  (9em,17.668em) {$\lambda$};
\draw (pc) -- (b);
\node (mid) at (9em,20.2em) {$\apsnodei$};
\draw (pc) -- (mid);
\node [tns] (tt) at (9em,22.468em) {};
\node (ttl) at (9em,22.468em) {$+$};
\draw (mid) -- (tt);
\node (left) at (6em,24.8em) {$\apsnodei$}; 
\node (leftt) at (6em,25.2em) {};
\node (right) at (12em,24.8em) {$\apsnodei$}; 
\draw (tt) -- (left);
\draw (tt) -- (right);
\draw plot [smooth, tension=1] coordinates { (leftt) (13em,27em) (13em,17em) (pc.south east)};
\node (sleeps) at (12em,25.5em) {\textit{sleeps}};
\node [tns] (pl) at (0em,6.932em) {};
\draw (pl) -- (ev);
\node (l) at  (0em,6.932em) {$\lambda$};
\node (e) at (0em,9.6em) {$\apsnodei$};
\draw (pl) -- (e);
\node (a) at (3em,14.4em) {$\apsnodei$};
\node (b) at (-3em,14.4em) {$\apsnodei$};
\node (p) at (-3em,14.8em) {};
\node[tns] (c) at (0em,12.268em) {}; 
\node (cl) at (0em,12.268em) {$@$}; 
\draw (c) -- (e);
\draw (c) -- (a);
\draw (c) -- (b);
\node (everyone) at (3em,15.1em) {\textit{everyone}};
\draw plot [smooth, tension=1] coordinates { (p) (4em,17.6em) (4em,7.6em) (pl.south east)};
\end{tikzpicture}
\begin{tikzpicture}[scale=0.75]
\node (ra) at (-6em,4.8em) {$\rightarrow$};
\node (raa) at (-6em,5.8em) {$[\beta]$};
\node (spar) at (2em,0em) {$s$};
\node (a) at (6em,4.8em) {$\apsnodei$};
\node (ev) at (-2em,4.8em) {$\apsnodei$};
\node[tns] (c) at (2em,2.668em) {}; 
\node (cl) at (2em,2.668em) {$@$}; 
\draw (c) -- (spar);
\draw (c) -- (a);
\draw (c) -- (ev);
%
\node [par] (pc) at (9em,6.932em) {};
\node (l) at  (9em,6.932em) {\nodeindex{$\lambda$}};
\node (e) at (9em,9.6em) {$\apsnodei$};
\path[>=latex,->]  (pc) edge (a);
\draw (pc) -- (e);
\node (a) at (12em,14.4em) {$\apsnodei$};
\node (leftt) at (6em,14.4em) {$\apsnodei$};
\node (lefttt) at (6em,15em) {};
\node[tns] (c) at (9em,12.268em) {}; 
\node (cl) at (9em,12.268em) {$+$}; 
\draw (c) -- (e);
\draw (c) -- (a);
\draw (c) -- (leftt);
\draw plot [smooth, tension=1] coordinates { (lefttt) (13em,17em) (13em,7em) (pc.south east)};
\node (sleeps) at (12em,15.5em) {\textit{sleeps}};
\node [tns] (pl) at (0em,6.932em) {};
\draw (pl) -- (ev);
\node (l) at  (0em,6.932em) {$\lambda$};
\node (e) at (0em,9.6em) {$\apsnodei$};
\draw (pl) -- (e);
\node (a) at (3em,14.4em) {$\apsnodei$};
\node (b) at (-3em,14.4em) {$\apsnodei$};
\node (p) at (-3em,14.8em) {};
\node[tns] (c) at (0em,12.268em) {}; 
\node (cl) at (0em,12.268em) {$@$}; 
\draw (c) -- (e);
\draw (c) -- (a);
\draw (c) -- (b);
%
%
\node (everyone) at (3em,15.1em) {\textit{everyone}};
\draw plot [smooth, tension=1] coordinates { (p) (4em,17.6em) (4em,7.6em) (pl.south east)};
\end{tikzpicture}
\begin{tikzpicture}[scale=0.75]
\node (ra) at (2em,4.8em) {$\rightarrow$};
\node (raa) at (2em,5.8em) {$[\beta]$};
\node (every) at (14em,5.5em) {\textit{everyone}};
\node (spar) at (10em,0em) {$s$};
\node (ev) at (14em,4.8em) {$\apsnodei$};
\node (lat) at (6em,4.8em) {$\apsnodei$};
\node[tns] (c) at (10em,2.668em) {}; 
\node (cl) at (10em,2.668em) {$@$}; 
\draw (c) -- (spar);
\draw (c) -- (lat);
\draw (c) -- (ev);
\node (a) at (12em,14.4em) {$\apsnodei$};
\node (leftt) at (6em,14.4em) {$\apsnodei$};
\node (lefttt) at (6em,15em) {};
\node[tns] (cc) at (9em,12.268em) {}; 
\node (cl) at (9em,12.268em) {$+$}; 
\draw (cc) -- (a);
\draw (cc) -- (leftt);
%
\node [par] (pc) at (9em,6.932em) {};
\node (l) at  (9em,6.932em) {\nodeindex{$\lambda$}};
\node (e) at (9em,9.6em) {$\apsnodei$};
\path[>=latex,->]  (pc) edge (lat);
\draw plot [smooth, tension=1] coordinates { (lefttt) (13em,17em) (12em,8em) (pc.south east)};
\node (sleeps) at (12em,15.5em) {\textit{sleeps}};
\draw (e) -- (pc);
\draw (e) -- (cc);
\end{tikzpicture}
\begin{tikzpicture}[scale=0.75]
\node (ra) at (2em,4.8em) {$\rightarrow$};
\node (raa) at (2em,5.8em) {$[\himpl I]$};
\node (every) at (14em,5.5em) {\textit{everyone}};
\node (spar) at (10em,0em) {$s$};
\node (ev) at (14em,4.8em) {$\apsnodei$};
\node (lat) at (6em,4.8em) {$\apsnodei$};
\node[tns] (c) at (10em,2.668em) {}; 
\node (cl) at (10em,2.668em) {$@$}; 
\draw (c) -- (spar);
\draw (c) -- (lat);
\draw (c) -- (ev);
\node (a) at (12em,14.4em) {$\apsnodei$};
\node (leftt) at (6em,14.4em) {$\apsnodei$};
\node (lefttt) at (6em,15em) {};
\node[tns] (cc) at (9em,12.268em) {}; 
\node (cl) at (9em,12.268em) {$+$}; 
\draw (cc) -- (a);
\draw (cc) -- (leftt);
%
\node [tns] (pc) at (9em,6.932em) {};
\node (l) at  (9em,6.932em) {$\lambda$};
\node (e) at (9em,9.6em) {$\apsnodei$};
\path[>=latex,->]  (pc) edge (lat);
\draw plot [smooth, tension=1] coordinates { (lefttt) (13em,17em) (12em,8em) (pc.south east)};
\node (sleeps) at (12em,15.5em) {\textit{sleeps}};
\draw (e) -- (pc);
\draw (e) -- (cc);
\end{tikzpicture}
\hspace{1ex}
\begin{tikzpicture}[scale=0.75]
  \node (ra) at (1em,24.8em) {$\rightarrow$};
\node (raa) at (1em,25.8em) {$[\beta]$};
\node (mid) at (9em,20.0em) {$s$};
%
\node [tns] (tt) at (9em,22.468em) {};
\node (ttl) at (9em,22.468em) {$+$};
\draw (mid) -- (tt);
\node (left) at (6em,24.8em) {$\apsnodei$}; 
\node (leftt) at (6em,25.2em) {};
\node (right) at (12em,24.8em) {$\apsnodei$}; 
\draw (tt) -- (left);
\draw (tt) -- (right);
\node (everyone) at (6em,25.5em) {\textit{everyone}};
\node (sleeps) at (12em,25.5em) {\textit{sleeps}};
\end{tikzpicture}
\end{center}
\caption{Conversion for the abstract proof structure corresponding to the proof structure of Figure~\ref{fig:expsb}}
\label{fig:exconv}
\end{figure}
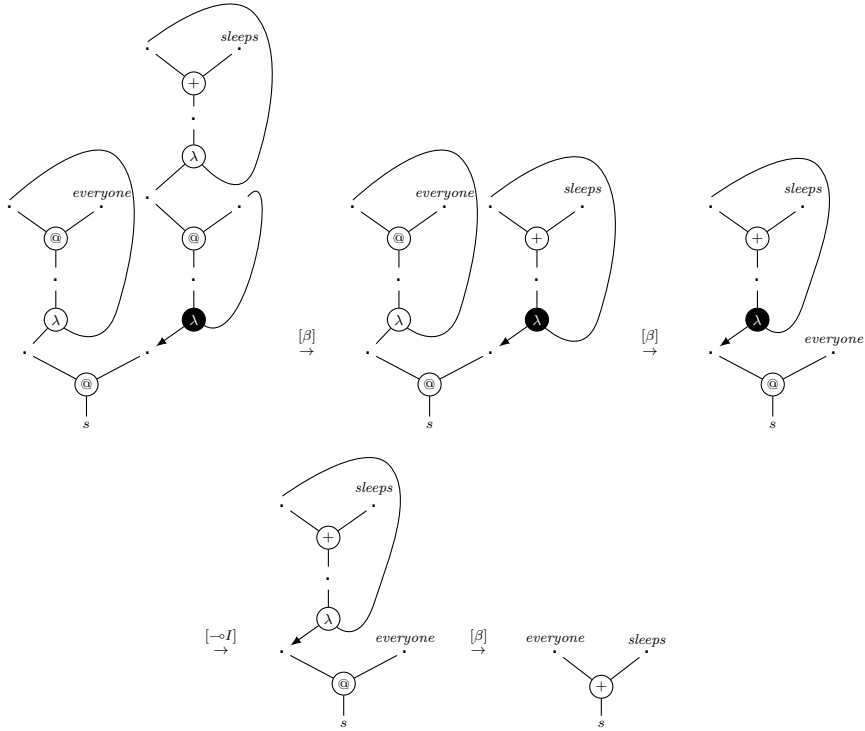

Figure~\ref{fig:exconv} shows how the abstract proof structure shown on the right hand side of Figure~\ref{fig:expsb} back on page~\pageref{fig:expsb} converts to the lambda graph $\textit{everyone}+\textit{sleeps}$ --- after one application of the $[\himpl I]$ conversion and three applications of the $\beta$ conversion --- and is therefore a proof net.

We show that a proof net with premisses $A_1,\ldots A_k$ and conclusion $C$ converts to a lambda graph $M$ whenever $x_1:A_1,\ldots,x_n:A_n\vdash M:C$ is derivable, and vice versa. 

\begin{lemma} If $\delta$ is a natural deduction proof of $x_1:A_1,\ldots,x_n:A_n\vdash M:C$, then we can construct a proof net with premisses $A_1,\ldots,A_n$ and conclusion $C$ contracting to $M$.
\end{lemma}

\paragraph{Proof} Induction on the length $l$ of $\delta$. If $l=0$, then we have either an axiom $x:A \vdash x:A$ or a lexicon rule $p:w \vdash M:A$ (with $M$ a linear lambda term with a single free variable $p$). In either case, the abstract proof structure will convert in zero steps to the required graph: the single vertex assigned logical hypothesis $x$ for the axiom rule, or the graph of $M$ with lexical hypothesis $p$ for the lexicon rule. 

If $l>0$, we look at the last rule of the proof and proceed by case analysis. 

If the last rule of the proof is the $/ I$ rule, we are in the follow case.

\[
\infer[/ I]{\Gamma \nd N:A/B}{\infer*[\delta]{\Gamma, x:B \nd N + x:A}{}}
\]

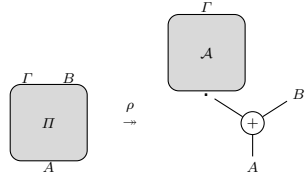
\begin{figure}
\begin{center}	
\begin{tikzpicture}[scale=0.75]
\node[pn] at (-10.3em,7.7em) {$\Pi$}; 
\node (ab) at (-10.3em,4.8em) {$A$};
\node (b) at (-9em,10.6em) {$B$};
\node (ab) at (3em,4.8em) {$A$};
\node (a) at (0,9.6em) {$\apsnodei$};
\node (b) at (6em,9.6em) {$B$};
\node[tns] (c) at (3em,7.668em) {};
\node (cl) at (3em,7.668em) {$+$};
\draw (c) -- (ab);
\draw (c) -- (a);
\draw (c) -- (b);
\node[pn] at (0.0em,12.3em) {$\mathcal{A}$}; 
%
\node (gamma) at (0.0em,15.3em) {$\Gamma$};
\node (labl) at (-5em,8.7em) {$\rho$}; 
\node  at (-5em,7.7em) {$\twoheadrightarrow$}; 
%
%
\node (gam) at (-11.7em,10.6em) {$\Gamma$};
\end{tikzpicture}
\end{center}
\caption{Conversion sequence obtained by induction hypothesis for the premiss of the $[/\textit{I}]$ rule.}
\label{fig:soundldr}.
\end{figure}

Removing the last rule gives us the shorter proof $\delta$, and induction hypothesis gives us a proof net of $\Gamma, x:B \vdash N+ x:A$. In other words, induction hypothesis gives us a proof net of $\Gamma, B \vdash A$ such that the underlying abstract proof structure converts, using a reduction sequence $\rho$, to $N + x$, with $x$ corresponding to $B$, as shown schematically in Figure~\ref{fig:soundldr}.

We need to produce a proof net of $\Gamma\vdash N: A/B$. But this is done simply by
adding the $/I$ link to the proof net of the induction hypothesis and adding a final $/I$ reduction as shown in Figure~\ref{fig:soundldrb}.

\begin{figure}
\begin{center}	
\begin{tikzpicture}[scale=0.75]
\node[pn] at (-10.3em,7.7em) {$\Pi$}; 
\node (ab) at (-9em,4.8em) {$A$};
\node (b) at (-9em,10.6em) {$B$};
\node (pa) at (-12em,0em) {$A/B$};
\node (pad) at (-12em,0em) {\phantom{M}};
\node[par] (pc) at (-9em,1.732em) {};
\node at (-9em,1.732em) {\nodeindex{+}};
\draw (pc) -- (ab);
\path[>=latex,->]  (pc) edge (pad);
\draw (b) to [out=50,in=330] (pc);
\node (ab) at (3em,4.8em) {$\apsnodei$};
\node (a) at (0,9.6em) {$\apsnodei$};
\node (b) at (6em,9.6em) {$\apsnodei$};
\node[tns] (c) at (3em,7.668em) {};
\node (cl) at (3em,7.668em) {$+$};
\draw (c) -- (ab);
\draw (c) -- (a);
\draw (c) -- (b);
\node (pa) at (0,0) {$A/B$};
\node (pad) at (0em,0em) {\phantom{M}};
\node[par] (pc) at (3em,1.732em) {};
\node at (3em,1.732em) {\nodeindex{+}};
\draw (pc) -- (ab);
\path[>=latex,->]  (pc) edge (pad);
\draw (b) to [out=50,in=330] (pc);
\node[pn] at (0.0em,12.3em) {$\mathcal{A}$}; 
\node (gamma) at (0.0em,15.3em) {$\Gamma$};
\node (gam) at (18.0em,11em) {$\Gamma$};
\node[pn] at (18.0em,8.0em) {$\mathcal{A}$}; 
\node at (18em,5.0em) {$A/B$}; 
\node (labl) at (-4em,6.0em) {$\rho$}; 
\node  at (-4em,5.0em) {$\twoheadrightarrow$}; 
\node (labl) at (10em,6.0em) {$[\ldr I]$}; 
\node  at (10em,5.0em) {$\rightarrow$}; 
\node (gam) at (-11.7em,10.6em) {$\Gamma$};
\end{tikzpicture}
\end{center}
\caption{Conversion sequence of a proof net ending with a $[\ldr I]$ contraction}
\label{fig:soundldrb}.
\end{figure}


The cases for $\backslash I$ and $\himpl$ are similar, adding the corresponding link and conversion to the proof net obtained by induction hypothesis.

The cases for the elimination rules $/ E$, $\backslash E$ and $\himpl E$ simply combine the two proof nets obtained by induction hypothesis with the corresponding link.

If the final rule is the $\beta\eta$ rule or a structural rule, we simply add, respectively, the $\beta$ reduction and the corresponding structural conversion.\qedsym


\begin{lemma} Given a proof net $\Pi$ with premisses $A_1,\ldots A_n$ and conclusion $C$ converting to a lambda graph $M$, there is a natural deduction proof $x_1:A_1,\ldots,x_n:A_n \vdash M:C$.
\end{lemma}

We proceed by induction on the number of conversions $c$.

If $c=0$ there are no conversions. As a consequence, there are no par links in the proof net. We proceed by induction on the number of tensor links $t$ in the proof net. 

If $t = 0$, the proof net consists of a single formula $A$ and the abstract proof structure is either a single vertex $x$ (in the case of a hypothesis), corresponding to a proof $x:A \vdash x:A$ or a term $M$ corresponding to a lexical entry, corresponding to a proof $p:w \vdash M:A$.

If $ t > 0$, then we can remove any tensor link and obtain three disjoint tensor trees and therefore three different proof nets, each with strictly less than $t$ tensor links. By induction hypothesis, we can therefore assume that we have three proofs $\delta_1$, $\delta_2$ and $\delta_3$, and we need to show we can combine these into the required proof for the tensor link.

We only verify the case for the $[\himpl I]$ link, the other cases are similar. In this case, the tensor tree and corresponding lambda graph look as follows.

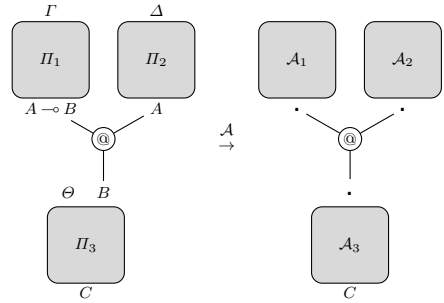
\begin{center}
 \begin{tikzpicture}[scale=0.75]
\node (lt) at (19em,-13.5em) {$\apsnodei$}; 
\node[tns] (tc) at (16em,-15.232em) {}; 
\node (appl) at (16em,-15.232em) {$@$};
\node (rt) at (13em,-13.5em) {$\apsnodei$}; 
\node (top) at (16em,-18.3em) {$\apsnodei$}; 
\draw (lt) -- (tc);
\draw (rt) -- (tc);
\draw (tc) -- (top);
\node[pn] at (13.0em,-10.7em) {$\mathcal{A}_1$}; 
\node[pn] at (19.0em,-10.7em) {$\mathcal{A}_2$}; 
\node[pn] at (16.0em,-21.3em) {$\mathcal{A}_3$}; 
\node (c) at (16em,-24.0em) {$C$};
\node at (9.0em,-15.7em) {$\rightarrow$}; 
\node at (9.0em,-14.7em) {$\mathcal{A}$}; 
\node (lt) at (5em,-13.5em) {$A$}; 
\node[tns] (tc) at (2em,-15.232em) {}; 
\node (appl) at (2em,-15.232em) {$@$};
\node (rt) at (-1em,-13.5em) {$A \himpl B$}; 
\node (top) at (2em,-18.3em) {$B$}; 
\draw (lt) -- (tc);
\draw (rt) -- (tc);
\draw (tc) -- (top);
\node[pn] at (-1em,-10.7em) {$\Pi_1$}; 
\node[pn] at (5.0em,-10.7em) {$\Pi_2$}; 
\node[pn] at (1.0em,-21.3em) {$\Pi_3$};
\node (gamma) at (-1em,-7.9em) {$\Gamma$};
\node (delta) at (5em,-7.9em) {$\Delta$};
\node (theta) at (0em,-18.3em) {$\Theta$};
\node (c) at (1em,-24.0em) {$C$};
\end{tikzpicture}
\end{center}

The three proofs we have by induction hypothesis are a proof $\delta_1$ of $\Gamma \vdash M:A\himpl B$ (with $M$ the term corresponding to the abstract proof structure $\mathcal{A}_1$), a proof $\delta_2$ of $\Delta \vdash N:A$ (with $N$ the term corresponding to the abstract proof structure $\mathcal{A}_2$) and a proof $\delta_3$ of $\Theta, x:B \vdash P[x]:C$ (with $P[]$ the context corresponding $\mathcal{A}_3$). We need to show we can combine these three proofs into a proof of $\Gamma,\Delta,\Theta \vdash P[(M\, N)]:C$, which we can do as follows (where $\textit{SL}$ again denotes application of the substitution lemma).

\[
\infer[\textit{SL}]{\Gamma,\Delta,\Theta\vdash P[(M\, N)]:C}{
  \infer*[\delta_3]{\Theta,x:B \vdash P[x]:C}{} &
\infer[\himpl E]{\Gamma,\Delta \vdash (M\, N):B}{
   \infer*[\delta_1]{\Gamma\vdash M:A\himpl B}{}
 & \infer*[\delta_2]{\Delta\vdash N:A}{}
}}
\]



If $c > 0$, we look at the last conversion and proceed by case ana\-ly\-sis.

If the last conversion is a $\beta\eta$ conversion or a structural rule, then induction hypothesis gives us a proof $\delta$ of $\Gamma\vdash M':C$, which we can extend using the $\beta$ rule (or structural rule) on $M'$ to produce $M$ and a proof of $\Gamma\vdash M:C$.

\begin{figure}
\begin{center}	
\begin{tikzpicture}[scale=0.75]
\node[pn] at (-10.3em,7.7em) {$\Pi_1$}; 
\node[pn] at (-13.0em,-3.0em) {$\Pi_2$}; 
\node (ab) at (-9em,4.8em) {$A$};
\node (b) at (-9em,10.6em) {$B$};
\node (pa) at (-12em,0em) {$A/B$};
\node (pad) at (-12em,0em) {\phantom{M}};
\node[par] (pc) at (-9em,1.732em) {};
\node at (-9em,1.732em) {\nodeindex{+}};
\draw (pc) -- (ab);
\path[>=latex,->]  (pc) edge (pad);
\draw (b) to [out=50,in=330] (pc);
\node at (-13.0em,-6.0em) {$C$};
\node (ab) at (3em,4.8em) {$\apsnodei$};
\node (a) at (0,9.6em) {$\apsnodei$};
\node (b) at (6em,9.6em) {$\apsnodei$};
\node[tns] (c) at (3em,7.668em) {};
\node (cl) at (3em,7.668em) {$+$};
\draw (c) -- (ab);
\draw (c) -- (a);
\draw (c) -- (b);
\node (pa) at (0,0) {$\apsnodei$};
\node[par] (pc) at (3em,1.732em) {};
\node at (3em,1.732em) {\nodeindex{+}};
\draw (pc) -- (ab);
\path[>=latex,->]  (pc) edge (pa);
\draw (b) to [out=50,in=330] (pc);
\node[pn] at (0.0em,12.3em) {$\mathcal{A}_1$}; 
\node[pn] at (0.0em,-2.7em) {$\mathcal{A}_2$}; 
\node at (0em,-5.7em) {$C$};
\node (delta) at (-1.2em,0.3em) {$\Delta$};
\node (gamma) at (0.0em,15.3em) {$\Gamma$};
\node (gam) at (18.0em,11em) {$\Gamma$};
\node (delta) at (14.8em,5.0em) {$\Delta$};
\node[pn] at (18.0em,8.0em) {$\mathcal{A}_1$}; 
\node[pn] at (16.0em,2.0em) {$\mathcal{A}_2$};
\node at (16em,-1em) {$C$};
\node at (18em,5.0em) {$\apsnodei$}; 
\node (labl) at (-4em,6.0em) {$\rho$}; 
\node  at (-4em,5.0em) {$\twoheadrightarrow$}; 
\node (labl) at (10em,6.0em) {$[\ldr I]$}; 
\node  at (10em,5.0em) {$\rightarrow$}; 
\node (gam) at (-11.7em,10.6em) {$\Gamma$};
\node (delta) at (-14.7em,0em) {$\Delta$};
\end{tikzpicture}
\end{center}
\caption{Conversion sequence of a proof net ending with a $[\ldr I]$ contraction}
\label{fig:completeldr}.
\end{figure}

If the last conversion is a $\ldr I$ contraction, we are in the case shown in Figure~\ref{fig:completeldr}.
The arrow marked by $\rho$ represents the conversion of the proof structure shown to its corresponding abstract proof structure, followed by any number of conversions $\rho$ to this abstract proof structure, with the final $[\ldr I]$ conversion indicated explicitly.

We need to show that we have a proof of $M[N]:C$, where $M[]$ is the term corresponding to the lambda graph $A_2$ (this is a term with a hole corresponding to the distinguished position indicated by a dot on the rightmost graph of Figure~\ref{fig:completeldr}) and $N$ is the lambda graph corresponding to $A_1$. Removing the $/I$ link and the final contraction produces the two structures shown in Figure~\ref{fig:completeldrb}

\begin{figure}
\begin{center}	
\begin{tikzpicture}[scale=0.75]
\node[pn] at (-10.3em,7.7em) {$\Pi_1$}; 
\node[pn] at (-13.0em,-3.0em) {$\Pi_2$}; 
\node (ab) at (-10.3em,4.8em) {$A$};
\node (b) at (-9em,10.6em) {$B$};
\node (pa) at (-12em,0em) {$A/B$};
\node at (-13.0em,-6.0em) {$C$};
\node (pa) at (1em,0em) {$A/B$};
\node (ab) at (3em,4.8em) {$A$};
\node (a) at (0,9.6em) {$\apsnodei$};
\node (b) at (6em,9.6em) {$B$};
\node[tns] (c) at (3em,7.668em) {};
\node (cl) at (3em,7.668em) {$+$};
\draw (c) -- (ab);
\draw (c) -- (a);
\draw (c) -- (b);
\node[pn] at (0.0em,12.3em) {$\mathcal{A}_1$}; 
\node[pn] at (0.0em,-3.0em) {$\mathcal{A}_2$}; 
\node at (0em,-6.0em) {$C$};
\node (delta) at (-1.2em,0.0em) {$\Delta$};
\node (gamma) at (0.0em,15.3em) {$\Gamma$};
\node (labl) at (-5em,8.7em) {$\rho_1$}; 
\node  at (-5em,7.7em) {$\twoheadrightarrow$}; 
\node (labl) at (-6.5em,-1.7em) {$\rho_2$}; 
\node  at (-6.5em,-2.7em) {$\twoheadrightarrow$}; 
\node (gam) at (-11.7em,10.6em) {$\Gamma$};
\node (delta) at (-14.7em,0em) {$\Delta$};
\end{tikzpicture}
\end{center}
\caption{Conversion sequence of Figure~\ref{fig:completeldr} with the final $[\ldr I]$ contraction removed.}
\label{fig:completeldrb}.
\end{figure}

All conversions in the conversion sequence $\rho$ are either fully in $\rho_1$ or fully in $\rho_2$, and $\rho_1$ and $\rho_2$ show that, respectively, $\Pi_1$ and $\Pi_2$ are proof nets. Given that either reduction sequence must be shorter than the reduction sequence of Figure~\ref{fig:completeldr} (because the final $[\ldr I]$ contraction has been removed), we can apply the induction hypothesis, 
giving us a proof $\delta_1$ of $\Gamma,x:B \vdash N+x:A$ and a proof $\Delta,z:A/B\vdash M[z]:C$. The term after the $[/I]$ contraction in the original proof net represents $M[N]$. We therefore need to create a proof of $\Gamma,\Delta\vdash M[N]:C$. This is done as follows.


\[
\infer[\textit{SL}]{\Gamma,\Delta\vdash M[N]:C}{
  \infer*[\delta_2]{\Delta,z:A/B\vdash M[z]:C}{}
  & \infer[/\textit{I}]{\Gamma\vdash N:A/B}{\infer*[\delta_1]{\Gamma,x:B\vdash N+x:A}{}}
  }
\]

The case for $\ldl$ is symmetric to the case for $\ldr$. 

If the last conversion is a $\himpl I$ contraction, we are in the case shown in Figure~\ref{fig:completehimpl}. The final lambda graph corresponds to the linear lambda term $M[\lambda x.N]$.

\begin{figure}
\begin{center}	
\begin{tikzpicture}[scale=0.75]
\node[pn] at (-10.3em,7.7em) {$\Pi_1$}; 
\node[pn] at (-13.0em,-3.0em) {$\Pi_2$}; 
\node (ab) at (-9em,4.8em) {$A$};
\node (b) at (-9em,10.6em) {$B$};
\node (pa) at (-12em,0em) {$B\himpl A$};
\node (pad) at (-12em,0em) {\rule{0em}{1.2em}};
\node[par] (pc) at (-9em,1.732em) {};
\node (pcl) at (-9em,1.732em) {\nodeindex{$\lambda$}};
\draw (pc) -- (ab);
\path[>=latex,->]  (pc) edge (pad);
\draw (b) to [out=50,in=330] (pc);
\node at (-13.0em,-6.0em) {$C$};
\node (gamma) at (0.3em,10.6em) {$\Gamma$};
\node (delta) at (-2.2em,0.3em) {$\Delta$};
\node (ab) at (3em,4.8em) {$\apsnodei$};
\node (b) at (3em,10.6em) {$\apsnodei$};
\node (pa) at (0,0) {$\apsnodei$};
\node[par] (pc) at (3em,1.732em) {};
\node (pcl) at (3em,1.732em) {\nodeindex{$\lambda$}};
\draw (pc) -- (ab);
\path[>=latex,->]  (pc) edge (pa);
\draw (b) to [out=50,in=330] (pc);
\node[pn] at (1.5em,7.7em) {$\mathcal{A}_1$}; 
\node[pn] at (-1.0em,-2.7em) {$\mathcal{A}_2$}; 
\node at (-1.0em,-5.7em) {$C$};
\node (gamma) at (18.2em,10.6em) {$\Gamma$};
\node (delta) at (15.8em,0.3em) {$\Delta$};
\node (ab) at (21em,4.8em) {$\apsnodei$};
\node (b) at (21em,10.6em) {$\apsnodei$};
\node (pa) at (18em,0) {$\apsnodei$};
\node[tns] (pc) at (21em,1.732em) {};
\node (pcl) at (21em,1.732em) {$\lambda$};
\draw (pc) -- (ab);
\draw (pc) -- (pa);
\draw (b) to [out=50,in=330] (pc);
\node[pn] at (19.5em,7.7em) {$\mathcal{A}_1$}; 
\node[pn] at (17.0em,-2.7em) {$\mathcal{A}_2$}; 
\node at (17em,-5.7em) {$C$};
\node (labl) at (10em,6.0em) {$[\himpl I]$}; 
\node  at (10em,5.0em) {$\rightarrow$}; 
\node (labl) at (-4em,6.0em) {$\rho$}; 
\node  at (-4em,5.0em) {$\twoheadrightarrow$}; 
\node (gam) at (-11.7em,10.6em) {$\Gamma$};
\node (delta) at (-14.7em,0em) {$\Delta$};
\end{tikzpicture}
\end{center}
\caption{Conversion sequence of a proof net ending with a $[\himpl I]$ contraction}
\label{fig:completehimpl}.
\end{figure}

Removing the final $[\himpl I]$ conversion produces two proof nets $\Pi_1$ and $\Pi_2$ with strictly shorter conversion sequences (again because we removed the final conversion and divided the other conversions) shown in Figure~\ref{fig:himplb}. Therefore, induction hypothesis gives us a proof $\delta_1$ of $\Gamma,x:B\vdash N:A$ and a proof $\delta_2$ of $\Delta,z:B\himpl A\vdash M[z]:C$ and we need to combine these into a proof of $\Gamma,\Delta\vdash M[\lambda x.N]:C$. This is done as follows.

\begin{figure}
\begin{center}	
\begin{tikzpicture}[scale=0.75]
\node[pn] at (-10.3em,7.7em) {$\Pi_1$}; 
\node[pn] at (-13.0em,-3.0em) {$\Pi_2$}; 
\node (ab) at (-9em,4.8em) {$A$};
\node (b) at (-9em,10.6em) {$B$};
\node (gam) at (-11.7em,10.6em) {$\Gamma$};
\node (delta) at (-14.7em,0em) {$\Delta$};
\node (pa) at (-12em,0em) {$B\himpl A$};
\node at (-13.0em,-6.0em) {$C$};
\node (ab) at (3em,4.8em) {$A$};
\node (b) at (3em,10.6em) {$B$};
\node (gam) at (0.3em,10.6em) {$\Gamma$};
\node (pa) at (1em,0em) {$B\himpl A$};
\node (delta) at (-1.7em,0em) {$\Delta$};
%
\node[pn] at (1.5em,7.7em) {$\mathcal{A}_1$}; 
\node[pn] at (0.0em,-3.0em) {$\mathcal{A}_2$}; 
\node at (0em,-6.0em) {$C$};
%
\node (labl) at (-5em,8.7em) {$\rho_1$}; 
\node  at (-5em,7.7em) {$\twoheadrightarrow$}; 
\node (labl) at (-6.5em,-1.7em) {$\rho_2$}; 
\node  at (-6.5em,-2.7em) {$\twoheadrightarrow$}; 
\end{tikzpicture}
\end{center}
\caption{Conversion sequence of Figure~\ref{fig:completehimpl} with the final $[\himpl I]$ conversion removed}
\label{fig:himplb}
\end{figure}


\[
\infer[\textit{SL}]{\Gamma,\Delta\vdash M[N]:C}{
  \infer*[\delta_2]{\Delta,z:B\himpl A\vdash M[z]:C}{}
  & \infer[\himpl\textit{I}]{\Gamma\vdash \lambda x. N:B \himpl A}{\infer*[\delta_1]{\Gamma,x:B\vdash N:A}{}}
  }
\]

%

\qedsym

\subsection{Cut elimination}

A standard sanity check for any logic is to check whether it satisfies cut elimination. We proved cut elimination for the sequent calculus in Section~\ref{sec:seqcutelim}. In a proof net, a cut formula is a formula which is the main formula of two links. By definition, a cut formula must then be a complex formula: an atomic formula must be an axiomatic formula.

Par links function as a kind of barrier in a proof net. Given a conversion sequence $\rho$ and a proof net $P$ with components $C$, we construct a tree $t$ as follows.
\begin{enumerate}
\item the leaves of the trees are the components,
\item for each structural conversion, add a unary branch to the component it operates on, 
\item for each par contraction we create a binary branch combining the two components the contraction is connected to into a new one.	
\end{enumerate}
Given this tree $t$, each structural rule operates inside a single component, and the initial conversion sequence $\rho$ is just one way of linearising $t$. Now, given a specific par link $p$, that is, a node in $t$, we can provide a conversion sequence $\rho'$ such that all conversions up until the contraction of the par link $p$ occur before all other conversions.



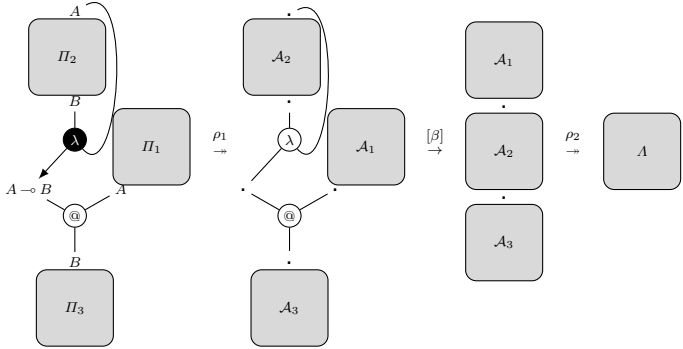
\begin{figure}
\begin{center}
\begin{tikzpicture}[scale=0.75]
\node[pn] at (39em,-11em) {$\Lambda$};
\node at (34.5em,-11em) {$\twoheadrightarrow$}; 
\node at (34.5em,-10em) {$\rho_2$}; 
\node[pn] at (30em,-11em) {$\mathcal{A}_2$}; 
\node[pn] at (30em,-5.0em) {$\mathcal{A}_1$}; 
\node[pn] at (30em,-17.0em) {$\mathcal{A}_3$}; 
\node at (30em,-8em) {$\apsnodei$};
\node at (30em,-14em) {$\apsnodei$}; 
\node at (25.5em,-11em) {$\rightarrow$}; 
\node at (25.5em,-10em) {$[\beta]$}; 
%
\node (lt) at (19em,-13.5em) {$\apsnodei$}; 
\node[tns] (tc) at (16em,-15.232em) {}; 
\node (appl) at (16em,-15.232em) {$@$};
\node (rt) at (13em,-13.5em) {$\apsnodei$}; 
\node (top) at (16em,-18.3em) {$\apsnodei$}; 
\draw (lt) -- (tc);
\draw (rt) -- (tc);
\draw (tc) -- (top);
\node[tns] (lam) at (16em,-10.232em) {}; 
\node (lamlab) at (16em,-10.232em) {$\lambda$}; 
\node (bl) at (16em,-7.7em) {$\apsnodei$}; 
\draw (bl) -- (lam);
\draw (lam) -- (rt);
\node[pn] at (15.5em,-4.8em) {$\mathcal{A}_2$}; 
\node[pn] at (21.0em,-10.7em) {$\mathcal{A}_1$}; 
\node[pn] at (16.0em,-21.3em) {$\mathcal{A}_3$}; 
\node (bot) at (16em,-2.0em) {$\apsnodei$}; 
\draw (lam) to [out=-50,in=-330] (bot); 
\node at (11.5em,-11em) {$\twoheadrightarrow$}; 
\node at (11.5em,-10em) {$\rho_1$}; 
\node (lt) at (5em,-13.5em) {$A$}; 
\node[tns] (tc) at (2em,-15.232em) {}; 
\node (appl) at (2em,-15.232em) {$@$};
\node (rt) at (-1em,-13.5em) {$A \himpl B$}; 
\node (top) at (2em,-18.3em) {$B$}; 
\draw (lt) -- (tc);
\draw (rt) -- (tc);
\draw (tc) -- (top);
\node[par] (lam) at (2em,-10.232em) {}; 
\node (lamlab) at (2em,-10.232em) {\nodeindex{$\lambda$}}; 
\node (bl) at (2em,-7.7em) {$B$}; 
\draw (bl) -- (lam);
\path[>=latex,->]  (lam) edge (rt);
\node[pn] at (1.5em,-4.8em) {$\Pi_2$}; 
\node[pn] at (7.0em,-10.7em) {$\Pi_1$}; 
\node[pn] at (2.0em,-21.3em) {$\Pi_3$}; 
\node (bot) at (2em,-1.8em) {$A$}; 
\draw (lam) to [out=-50,in=-330] (bot); 
\end{tikzpicture}
\end{center}
\caption{Conversion sequence for a cut formula with main connective `$\himpl$'.}
\label{fig:cuthimpl}
\end{figure}

\citet{mp} use this property to prove cut elimination for multimodal proof nets and their method can be adapted to the current context without problem. Only the case for $\himpl$ requires some thought. If we have a cut formula of the form $A\himpl B$, then we are in the situation shown in Figure~\ref{fig:cuthimpl}. A cut on a formula $A\himpl B$ looks as shown on the left. We reorder the conversion such that all conversions on the abstract proof structure $\mathcal{A}(\Pi_2)$ are in the initial sequence $\rho_1$, with the $\lambda$ rule as the last rule of $\rho_1$. We then apply the $\beta$ rule followed by the other conversions (first those outside of $\mathcal{A}(\Pi_2)$, then those occurring after the $\beta$ rule).

We can now simply reconnect the structures $\Pi_1$, $\Pi_2$ and $\Pi_3$ as shown in Figure~\ref{fig:cuthimplb}. We set $\rho'_1$ to be the sequence $\rho_1$ without the final $\lambda$ rule, keep $\rho_2$ unchanged and then the sequence in Figure~\ref{fig:cuthimplb} shown we obtain a proof net with the same structure $\Lambda$.

\begin{figure}
\begin{center}
\begin{tikzpicture}[scale=0.75]
\node[pn] at (39em,-11em) {$\Lambda$};
\node at (34.5em,-11em) {$\twoheadrightarrow$}; 
\node at (34.5em,-10em) {$\rho_2$}; 
\node[pn] at (30em,-11em) {$\mathcal{A}_2$}; 
\node[pn] at (30em,-5.0em) {$\mathcal{A}_1$}; 
\node[pn] at (30em,-17.0em) {$\mathcal{A}_3$}; 
\node at (30em,-8em) {$\apsnodei$};
\node at (30em,-14em) {$\apsnodei$}; 
\node at (25.5em,-11em) {$\twoheadrightarrow$}; 
\node at (25.5em,-10em) {$\rho_1$}; 
\node[pn] at (21em,-11em) {$\Pi_2$}; 
\node[pn] at (21em,-5.0em) {$\Pi_1$}; 
\node[pn] at (21em,-17.0em) {$\Pi_3$}; 
\node at (21em,-8em) {$A$};
\node at (21em,-14em) {$B$}; 
\end{tikzpicture}
\end{center}
\caption{Replacing the cut formula for `$\himpl$' by two simpler cuts.}
\label{fig:cuthimplb}
\end{figure}

\subsection{Eager application of the contractions}

Given confluence of the rewrite operations, we can apply partial evaluation to lexical abstract proof structures. This is similar to partial evaluation of the semantics
as done by \citet{m99geo} and by \citet{sempn}.

For example, the abstract proof structure for `sleeps' is shown below on the left. Since we are in the correct configuration for a lambda conversion (because we require
lexical entries to be assigned linear lambda terms, we can always directly apply lambda conversions to terms coming from the lexicon), this produces the structure shown
in the middle. However, this is a valid $\beta$ redex and reducing it produces the structure shown below on the right. This structure is identical to directly unfolding a
lexical entry `sleeps' assigned  $np\backslash s$. 

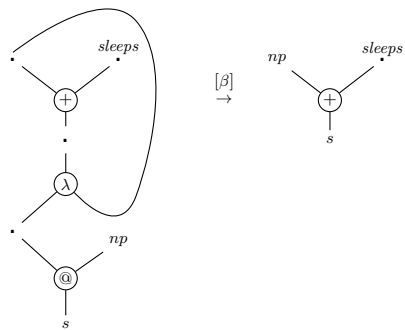
\begin{center}
\begin{tikzpicture}[scale=0.75]
\node (e) at (24em,9.6em) {$s$};
\node (a) at (27em,14.4em) {$np$};
\node (b) at (21em,15em) {$\apsnodei$};
\node[tns] (c) at (24em,12.268em) {}; 
\node (cl) at (24em,12.268em) {$@$}; 
\draw (c) -- (e);
\draw (c) -- (a);
\draw (c) -- (b);
\node [tns] (pc) at (24em,17.668em) {};
\node (l) at  (24em,17.668em) {$\lambda$};
\draw  (pc) -- (b);
\node (mid) at (24em,20.2em) {$\apsnodei$};
\draw (pc) -- (mid);
\node [tns] (tt) at (24em,22.468em) {};
\node (ttl) at (24em,22.468em) {$+$};
\draw (mid) -- (tt);
\node (left) at (21em,24.8em) {$\apsnodei$}; 
\node (leftt) at (21em,25.2em) {};
\node (right) at (27em,24.8em) {$\apsnodei$}; 
\draw (tt) -- (left);
\draw (tt) -- (right);
\draw plot [smooth, tension=1] coordinates { (leftt) (28em,27em) (28em,17em) (pc.south east)};
\node (sleeps) at (27em,25.5em) {\textit{sleeps}};
\node (mid) at (39em,20.2em) {$s$};
\node [tns] (tt) at (39em,22.468em) {};
\node (ttl) at (39em,22.468em) {$+$};
\draw (mid) -- (tt);
\node (left) at (36em,24.8em) {$np$}; 
\node (leftt) at (36em,25.2em) {};
\node (right) at (42em,24.8em) {$\apsnodei$}; 
\node (rightl) at (42em,25.4em) {\textit{sleeps}}; 
\draw (tt) -- (left);
\draw (tt) -- (right);
\node (labl) at (33em,23.5em) {$[\beta]$}; 
\node  at (33em,22.5em) {$\rightarrow$}; 
\end{tikzpicture}
\end{center}

Similarly, we can unfold the lexicon entry for everyone, then apply the beta reduction to obtain the simpler form shown below on the right. 

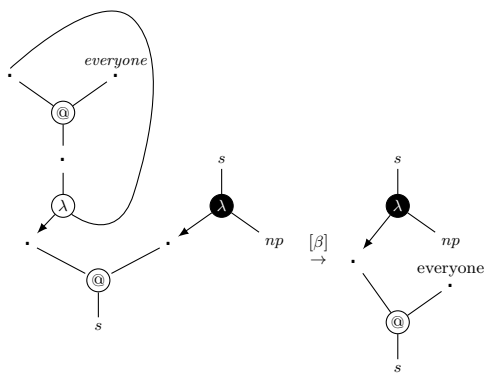
\begin{center}
\begin{tikzpicture}[scale=0.75]
\node (spar) at (2em,0em) {$s$};
\node (a) at (6em,4.8em) {$\apsnodei$};
\node (ev) at (-2em,4.8em) {$\apsnodei$};
\node[tns] (c) at (2em,2.668em) {}; 
\node (c) at (2em,2.668em) {$@$}; 
\draw (c) -- (spar);
\draw (c) -- (a);
\draw (c) -- (ev);
%
\node [par] (pc) at (9em,6.932em) {};
\node (l) at  (9em,6.932em) {\nodeindex{$\lambda$}};
\node (e) at (9em,9.6em) {$s$};
\node (f) at (12em,4.8em) {$np$}; 
\path[>=latex,->]  (pc) edge (a);
\draw (pc) -- (f);
\draw (pc) -- (e);
%
%
\node [tns] (pl) at (0em,6.932em) {};
\path[>=latex,->]  (pl) edge (ev);
\node (l) at  (0em,6.932em) {$\lambda$};
\node (e) at (0em,9.6em) {$\apsnodei$}; 
\draw (pl) -- (e);
%
\node (a) at (3em,14.4em) {$\apsnodei$};
\node (b) at (-3em,14.4em) {$\apsnodei$};
\node (p) at (-3em,14.8em) {};
\node[tns] (c) at (0em,12.268em) {}; 
\node (cl) at (0em,12.268em) {$@$}; 
\draw (c) -- (e);
\draw (c) -- (a);
\draw (c) -- (b);
%
%
\node (everyone) at (3em,15.1em) {\textit{everyone}};
\draw plot [smooth, tension=1] coordinates { (p) (4em,17.6em) (4em,7.6em) (pl.south east)};
\node (b) at (16.5em,3.8em) {$\apsnodei$};
\node [par] (pc) at (19em,6.932em) {};
\node (l) at  (19em,6.932em) {\nodeindex{$\lambda$}};
\node (e) at (19em,9.6em) {$s$};
\node (f) at (22em,4.8em) {$np$}; 
\path[>=latex,->]  (pc) edge (b);
\draw (pc) -- (f);
\draw (pc) -- (e);
\node (e) at (19em,-2.4em) {$s$};
\node (a) at (22em,2.4em) {$\apsnodei$};
\node (ev) at (22em,3.2em) {everyone};
\node[tns] (c) at (19em,0.268em) {}; 
\node (cl) at (19em,0.268em) {$@$}; 
\draw (c) -- (e);
\draw (c) -- (a);
\draw (c) -- (b);
\node (labl) at (14.5em,4.8em) {$[\beta]$}; 
\node  at (14.5em,3.8em) {$\rightarrow$}; 
\end{tikzpicture}
\end{center}

\section{Complexity}
\label{sec:compl}

Given the proof net calculus described in the previous sections, complexity analysis of hybrid type-logical grammars and several of its variants becomes simple.

\begin{proposition}\label{prop:np} If we can show the contraction criterion for proof structures can be performed in polynomial time, then deciding provability for the corresponding grammar logic is in NP.
\end{proposition}

\paragraph{Proof} To show that a problem is in NP, we only need to show that we can verify that a candidate solution is an actual solution in polynomial time. We can non-deterministically select a formula from the lexicon for each word, then non-deterministically enumerate all proof structures for these choices. This enumerates all candidate solutions. Since we have assumed polynomial time contractability of the correctness condition for proof structures, this means the logic is in NP.  \qedsym

\begin{theorem}\label{thm:np} HTLG parsing is NP complete

\end{theorem} 

\paragraph{Proof} Since HTLG contains lexicalized ACG as a fragment, NP-hardness follows from Proposition~5 of \citet{yk05acg}, so all that remains to be shown is that HTLG is in NP.

Since the contraction criterion can be verified in polynomial time (each contraction reduces the size of the proof structure, so verifying correctness is at most quadratic in the size of the proof structure),  Proposition~\ref{prop:np} applies. Therefore HTLG parsing is NP complete. 
 \qedsym

The proof of Theorem~\ref{thm:np} is very general and can easily be adapted to variants and extensions of HTLG. For example, we can add the connectives for `$\bullet$', `$\Diamond$' and `$\Box$' and mode information (as in the multimodal versions of the Lambek calculus \cite{M95}) while maintaining NP-completeness. 

When adding structural rules, complexity analysis becomes more delicate. Adding associativity, as in the original formulation of hybrid type-logical grammars, doesn't change the complexity, since we can simply use the strategy discussed in Section~\ref{sec:ass} to ensure polynomial contraction of proof structures. So we can actually strengthen Theorem~\ref{thm:np} to the following.
 
 \begin{theorem}\label{thm:npbis} HTLG$_{/_i,\bullet_i,\backslash_i,\Diamond_i,\Box_i}$ parsing, with associativity for some modes $i$, is NP complete.
\end{theorem} 

In general, NP completeness will be preserved whenever we provide the set of structural rules with a polynomial time contraction algorithm. When we do not have a polynomial contraction algorithm, we still have information about the complexity class:  when we add structural rules but use the standard restriction that the tree rewrites allowed by the structural rules are linear (no copying or deletion of leaves) and do not increase the size of the tree, then the resulting logic is PSPACE complete, following the argument of \citet[Section~9.2]{diss}. 

\begin{theorem} HTLG$_{/_i,\bullet_i,\backslash_i,\Diamond_i,\Box_i}$ parsing with any finite set of non-expanding structural rules is PSPACE complete.
\end{theorem}

This gives an NP lower bound and a PSPACE upper bound for any HTLG augmented with the multimodal connectives and a fixed set of structural rules, and NP completeness can be shown by providing a polynomial contraction algorithm.

Adding the additive connectives similarly produces a PSPACE upper bound, using the same argument as \citet{l92decision}: using a non-deterministic Turing machine for cut-free proof search and exploiting the linear bound on the depth of the sequents\footnote{We need to prove cut elimination for the multiplicative-additive logic for this to hold, which is rather simple extension of the cut elimination proof in Section~\ref{sec:seqcutelim}.}. 
PSPACE hardness follows from a result by \citet{kks19malc}, who show that the Lambek calculus with one slash extended with one additive connective (either additive disjunction or additive conjunction) is PSPACE complete. Therefore HTLG, even with the addition of only one of the additives, is PSPACE complete.


\begin{theorem} Multiplicative-additive HTLG is PSPACE complete.
\end{theorem}

Finally, given that the Lambek calculus with assignments to the empty string is known to be undecidable \cite{b82}, adding assignments to the empty string to HTLG results in an undecidable system as well. Similarly, since the Lambek calculus with second-order quantifiers is known to be undecidable \cite{emms95}, the same holds for HTLG. We leave open the question of whether the methods for making second-order quantifiers decidable in the categorial grammar and linear logic tradition \cite{emms,moortgat,perrier} can be adapted to type-logical grammar.

\begin{theorem}
HTLG with assignments to the empty string is undecidable.	
\end{theorem}

\begin{theorem}
HTLG with second-order quantification is undecidable.	
\end{theorem}
 
\section{Conclusion}

We have investigated the formal properties of hybrid type-logical grammars. Notably we proved some standard properties of the natural deduction calculus of the logic and provided two alternative (but equivalent) logical calculi for hybrid type-logical grammars: sequent calculus and proof nets. Although these alternative calculi are relatively standard extensions of such calculi for other type-logical grammars, presenting them for HTLG adds new tools for the formal study of HTLG.

These results help us better understand the theoretical foundations of the system, a question left open by \citet{kl13dgap}.

For future work, the complexity results of Section~\ref{sec:compl} can probably be sharpened, for example for the multimodal version of HTLG presented by \citet{kl20tls}. We have not looked at  model theory for HTLG (although, indirectly, cut elimination entails the existence of a model). An interesting open question would be to find a combination of one of the standard models for the Lambek calculus with one of the standard models for the lambda calculus and prove this sound and complete for HTLG.


\bibliographystyle{spmpscinat}
\bibliography{moot}


\end{document}